\documentclass[letterpaper]{article} 
\usepackage{aaai2026}  
\nocopyright 

\usepackage{times}  
\usepackage{helvet}  
\usepackage{courier}  
\usepackage[hyphens]{url}  
\usepackage{graphicx} 
\usepackage{natbib}  
\usepackage{caption} 
\usepackage{amssymb} 
\usepackage{algorithm,algpseudocode}
\algrenewcommand\algorithmiccomment[1]{\hfill$\triangleright$~#1}
\algtext*{EndFor}
\algtext*{EndIf}
\usepackage{enumitem} 
\usepackage{booktabs}           
\usepackage{tcolorbox}
\tcbuselibrary{breakable}
\usepackage{amsmath}

\usepackage[table]{xcolor}
\definecolor{lampbg}{HTML}{EAF4FF}
\usepackage{newfloat}
\usepackage{listings}
\DeclareCaptionStyle{ruled}{labelfont=normalfont,labelsep=colon,strut=off} 
\floatstyle{ruled}
\newfloat{listing}{tb}{lst}{}
\floatname{listing}{Listing}
\title{Think, Speak, Decide: Language-Augmented Multi-Agent Reinforcement Learning for Economic Decision-Making}
\author{
    Heyang Ma\textsuperscript{\rm 1,\rm 2,\rm 3,\rm4}\equalcontrib,
    Qirui Mi\textsuperscript{\rm 1,\rm 5}\equalcontrib,
    Qipeng Yang\textsuperscript{\rm 6,\rm 2,\rm 3},
    Zijun Fan\textsuperscript{\rm 6, \rm 2,\rm 3},
    Bo Li\textsuperscript{\rm 7},
    Haifeng Zhang\textsuperscript{\rm 1,\rm 2,\rm 5}\thanks{Corresponding author: haifeng.zhang@ia.ac.cn}
}
\affiliations{
    \textsuperscript{\rm 1}Institute of Automation, Chinese Academy of Sciences,
    \textsuperscript{\rm 2}Nanjing Artificial Intelligence Research of IA\\
    \textsuperscript{\rm 3}University of Chinese Academy of Sciences, Nanjing, 
    \textsuperscript{\rm 4}University of International Business and Economics\\
    \textsuperscript{\rm 5}School of Artificial Intelligence, Chinese Academy of Sciences,
    \textsuperscript{\rm 6}Nanjing University of Posts and Telecommunications\\
    \textsuperscript{\rm 7}School of Economics, Peking University\\
    haifeng.zhang@ia.ac.cn
}

\begin{document}

\maketitle

\begin{abstract}
Economic decision‑making depends not only on structured signals—such as prices and taxes—but also on unstructured language, including peer dialogue and media narratives. While multi‑agent reinforcement learning (MARL) has shown promise in optimizing economic decisions, it struggles with the semantic ambiguity and contextual richness of language. We propose \textbf{LAMP} (\textbf{L}anguage‑\textbf{A}ugmented \textbf{M}ulti‑Agent \textbf{P}olicy), the framework to integrate language into economic decision‑making, narrowing the gap to real‑world settings.
LAMP follows a \textbf{Think–Speak–Decide} pipeline:
\textbf{(1) Think} interprets numerical observations to extract short‑term shocks and long‑term trends, caching high‑value reasoning trajectories.
\textbf{(2) Speak} crafts and exchanges strategic messages based on the reasoning, updating beliefs by parsing peer communications.
\textbf{(3) Decide} fuses numerical data, reasoning, and reflections into a MARL policy to optimize language‑augmented decision‑making.
Experiments in economic simulation show that LAMP outperforms both MARL and LLM‑only baselines in cumulative return (\textbf{+63.5\%, +34.0\%}), robustness (\textbf{+18.8\%, +59.4\%}), and interpretability. These results demonstrate the potential of language‑augmented policies to deliver more effective and robust economic strategies.
\end{abstract}

\begin{links}
    \link{Code}{https://github.com/hey0223/LAMP}
\end{links}

\section{Introduction}

Real-world economic settings are rich in multi-agent interactions and decision-making challenges, spanning labor markets, firm pricing, and government policy design. Solving these economic decision‑making problems can yield explanatory insights into economic phenomena and prescriptive guidance for policy and strategy design~\cite{tversky1974judgment,varian1992microeconomic}. However, their characteristics—dynamic interactions, long‑term incentives, and uncertainty-make them substantially more challenging than conventional fixed‑rule benchmarks with fully specified dynamics~\cite{charpentier2023reinforcement,mi2023taxai}.
Recent advances in artificial intelligence (AI), particularly reinforcement learning (RL), have been applied to model and optimize economic decision‑making processes, with applications spanning household savings~\cite{rui2022aiagenthitmoving, rui2022learningzeromakeconsumptionsaving, atashbar2023ai}, market pricing~\cite{danassisAIdrivenPricesExternalities2023}, and tax policy~\cite{zheng2022ai,mi2023taxai,mi2025learning}. These studies provide evidence that RL can effectively address dynamic, multi‑agent economic problems.

\begin{figure}
    \centering
\includegraphics[width=1\linewidth]{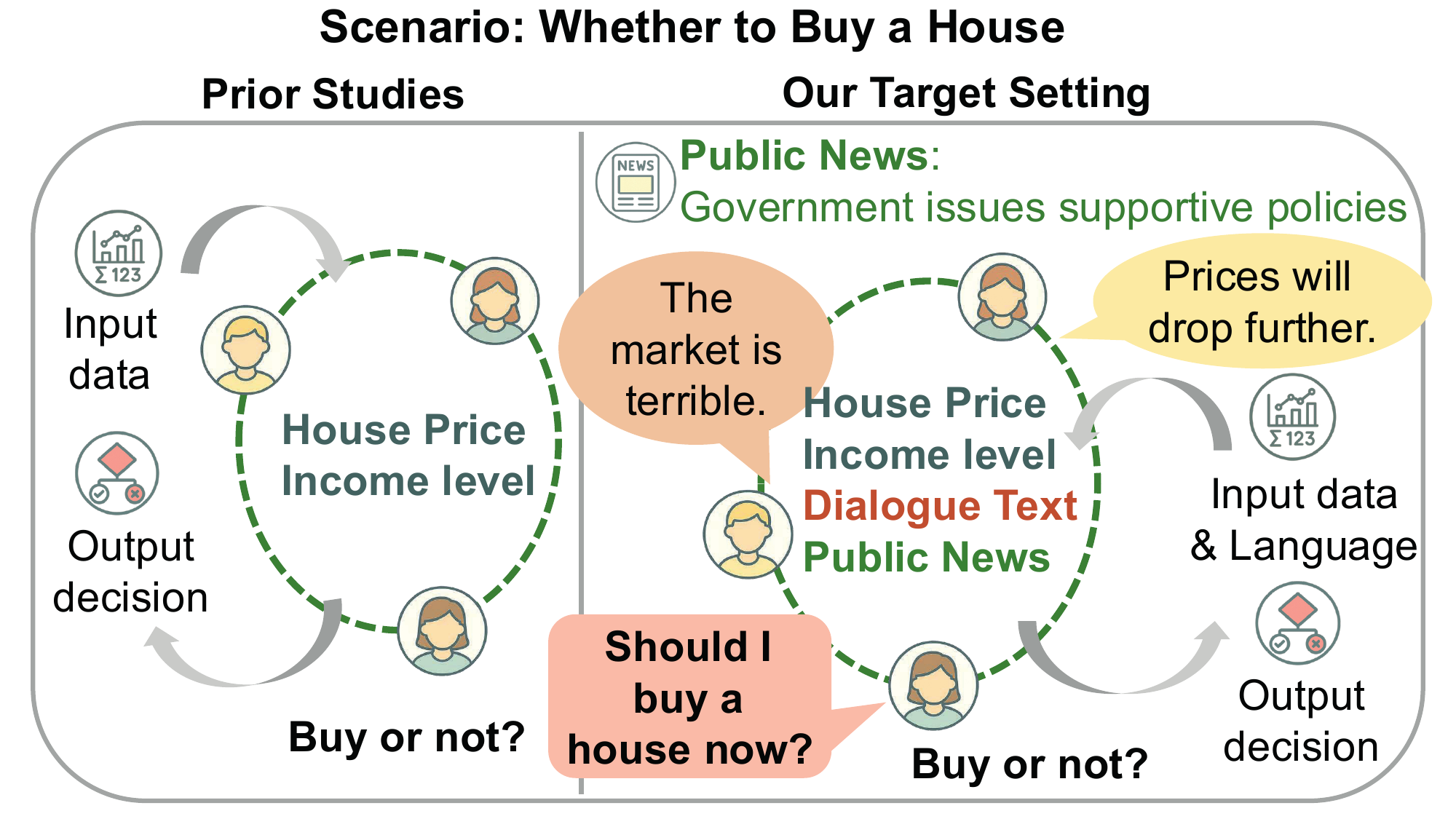}
\caption{Comparison of prior studies and our target: Unstructured language signals, alongside structured numerical data, are critical to economic decision‑making.}
  \label{fig1}
      
\end{figure}

However, economic decision‑making relies not only on numerical signals but also on language‑based information, such as peer dialogue and media narratives~\cite{luketina2019survey}. The above-mentioned RL-based studies largely ignore the impact of language. Standard MARL algorithms typically assume clean, structured communication protocols~\cite{zhu2024survey}, whereas real‑world economic decisions involve noisy, semantically rich, and sometimes deceptive natural language.
Large language models (LLMs) offer powerful tools to process such language. 
Recent work in policy evaluation~\cite{li2024econagent,hao2025multi}, trading~\cite{xiao2024tradingagents}, and simulation~\cite{mi2025econgym} demonstrates LLMs’ potential for language‑aware economic modeling. However, most employ LLMs to generate actions or simulate behaviors, without systematically optimizing agents’ policies. This remains insufficient for solving complex economic problems or producing robust, actionable policy insights. We therefore focus on the key question: \textbf{In complex multi‑agent economic environments, how can agents interpret and leverage natural‑language information to support optimal decisions?}

To address this, we propose \textbf{LAMP} (\textbf{L}anguage‑\textbf{A}ugmented \textbf{M}ulti‑\textbf{A}gent \textbf{P}olicy Learning), which integrates LLM‑driven reasoning and reflection over both numerical observations and textual signals to support optimal decision‑making. LAMP follows a unified \textbf{Think–Speak–Decide} pipeline:
(1) \textbf{Think}: Agents receive environment observations and generate both short‑term shock analysis and long‑term trend reasoning. High‑reward reasoning trajectories are stored in an experience pool for retrieval. The long‑term reasoning is also passed to the Speak module to inform message generation.
(2) \textbf{Speak}: Guided by the Think module, each agent formulates multiple candidate public messages. A lightweight attention‑based scorer selects one for broadcast. Other agents parse the message via the LLM, updating their beliefs, trust, and reflective states. These updated reflections are then passed to the Decide module.
(3) \textbf{Decide}: The policy network integrates numerical observations, \textit{Think}’s reasoning outputs, and \textit{Speak}’s reflections into the RL policy. Under centralized training with a shared critic, agents learn strategies capable of processing reasoning and reflection signals to produce robust, language‑aware economic decisions.
We evaluate LAMP in TaxAI and show that it outperforms MARL and LLM-only baselines in both returns and shock robustness. Its reasoning traces explain language-guided choices, aiding insight and policy. \textbf{Our contributions are threefold:}
\begin{enumerate}
\item \textbf{Framework}: We propose LAMP, a language‑augmented MARL framework that models the role of natural language in economic decision‑making, bringing it closer to real‑world contexts.
\item \textbf{Mechanism}: We introduce the \textit{Think–Speak–Decide} pipeline, explicitly structuring how agents reason over trends, exchange and interpret strategic messages, and integrate these insights into policy optimization.
\item \textbf{Empirical Results}: LAMP surpasses MARL and LLM‑only baselines in language‑guided decision performance, while providing interpretable reasoning trajectories for transparent policy analysis.
\end{enumerate}

\begin{algorithm}[h]
  \caption{Language-Augmented Multi-agent Policy}
  \label{alg:lamp-core}
  \small
  \begin{algorithmic}[1]
    \For{episode $e=1,2,\dots$}
      \State Reset environment; clear short experience
      \For{$t=0$ to $T$}
        \State Determine news type: $type \gets long,short,none$
        \State Generate news: $\mathcal{R}_t \gets \textbf{Think}(\cdot)_{type}$
        \ForAll{agents $i$}
             \State Clear the current step’s experience $\mathcal{H}_{k,t}^i$
            \If{$t$ is long‐term checkpoint}
              \State Retrieve $\mathcal{H}_{k,t}^i$ from $\mathcal{H}^{\text{long}}$ and $\mathcal{H}_{t,i}^{\text{short}}$
            \EndIf
            \State Generate economic status and reasoning:
            \State $\mathcal{L}_{\text{reason}}(\mathcal{R}_t,O_t^{h,i},\mathcal{H}_{k,t}^i)$ 
              \If{$t$ is long‐term checkpoint}
              \State Generate statement: $v_t^{i} \gets
                     \textbf{Speak}(O_t^{h,i},\mathcal{R}_t)$
              \State  Self-reflection and update belief and trust:
              \State $\left(w_{t}^{i\rightarrow j},\, \tau_{t}^{i\rightarrow j},\, \alpha_t^{i}\right) \leftarrow \mathcal{L}_{\mathrm{reflect}}(\cdot)$
            \EndIf
         \State Generate action: $a_t^{i} \gets
                     \mu_{\theta_i}(o_t^{i},
                     E_{\text{text}}(\cdot))$
        \EndFor
        \State Execute $a_t$; observe $(r_t,x_{t+1})$; store in $\mathcal{D}$
        \State Update $Q_\phi,\{\theta_i\}$ from $\mathcal{D}$
          \State Harvest top trajectories $\to$ short experience $\mathcal{H}_{t,i}^{\text{short}}$
      \EndFor
     \State Harvest top trajectories $\to$ long experience $\mathcal{H}^{\text{long}}$
    \EndFor
  \end{algorithmic}
\end{algorithm}

\section{Related Work}
\textbf{RL for Economic Decision-Making.}
Artificial intelligence provides a powerful computational tool for solving complex economic decision‑making problems. Early work includes Bayesian structural time series for policy causal inference~\cite{brodersen2015inferring} and heuristic search for tax design~\cite{malecka2020application}, but these approaches struggle with real-world complexity. Reinforcement learning (RL) now supports a broad macroeconomic research agenda, including tax policy design such as TaxAI~\cite{mi2023taxai}, monetary rule learning~\cite{chenDeepReinforcementLearning2023}, trade bargaining~\cite{sch2021intelligence}, learning-based heterogeneous-agent macroeconomic modeling~\cite{kurikshaEconomyNeuralNetworks2021}, and large‑population policy learning~\cite{zhao2024mean, mi2025learning}. At the microeconomic level, RL has modeled household consumption–saving behavior~\cite{rui2022aiagenthitmoving, rui2022learningzeromakeconsumptionsaving}, responses to income shocks~\cite{atashbar2023ai}, and emergent barter and exchange~\cite{johansonEmergentBarteringBehaviour2022, ozhamaratliDeepReinforcementLearning2022a}. 
While these studies show RL’s effectiveness in economic decision-making, they largely ignore language signals—policy debates, media reports, public opinion—thereby oversimplifying real-world settings.

\textbf{LLMs for Economic Research.}
 LLMs excel at processing language signals, and recent studies have explored their applications in economics. \emph{Homo Silicus} models human fairness and risk aversion~\cite{horton2023large}. \emph{Generative Agents} simulate sandbox societies~\cite{park2023generative}. \emph{EconAgent} uses LLM agents to evaluate fiscal and monetary policies~\cite{li2024econagent}. Other studies extend LLM agents to population behavior simulation~\cite{mi2025mf}, long‑term financial planning~\cite{douglas2024consumption}, and market trading~\cite{xiao2024tradingagents, yu2024fincon}. General platform \emph{EconGym}~\cite{mi2025econgym} benchmarks LLM agents in diverse economic scenarios. While these studies demonstrate the versatility of LLMs in economics, most remain focused on direct action generation or simulation, leaving open questions about their role in optimizing economic policies.

\textbf{Integration of MARL and LLMs.}  
We focus on combining MARL’s strength in policy optimization for multi‑agent settings with LLMs’ capacity to process language signals. Recent work explores this direction: \emph{FAMA} aligns LLM knowledge for multi‑agent coordination~\cite{slumbers2024leveraging}; \emph{LAMARL} uses LLM‑generated priors for policy and reward design~\cite{11027664}; \emph{MAPoRL} co‑trains LLMs to enhance cooperation~\cite{park2025maporl}; and \emph{CORY} fine‑tunes duplicated LLM agents in cooperative settings~\cite{ma2024cory}.  
Economic decision‑making is typically dynamic, non‑cooperative, and long‑horizon. Agents must interpret diverse numerical signals alongside semantically rich and potentially noisy language inputs, rendering prior MARL–LLM methods inadequate for such settings.

\section{Language-Augmented Multi-Agent Policy}

\begin{figure*}[t]
\centering
\includegraphics[width=1\textwidth]{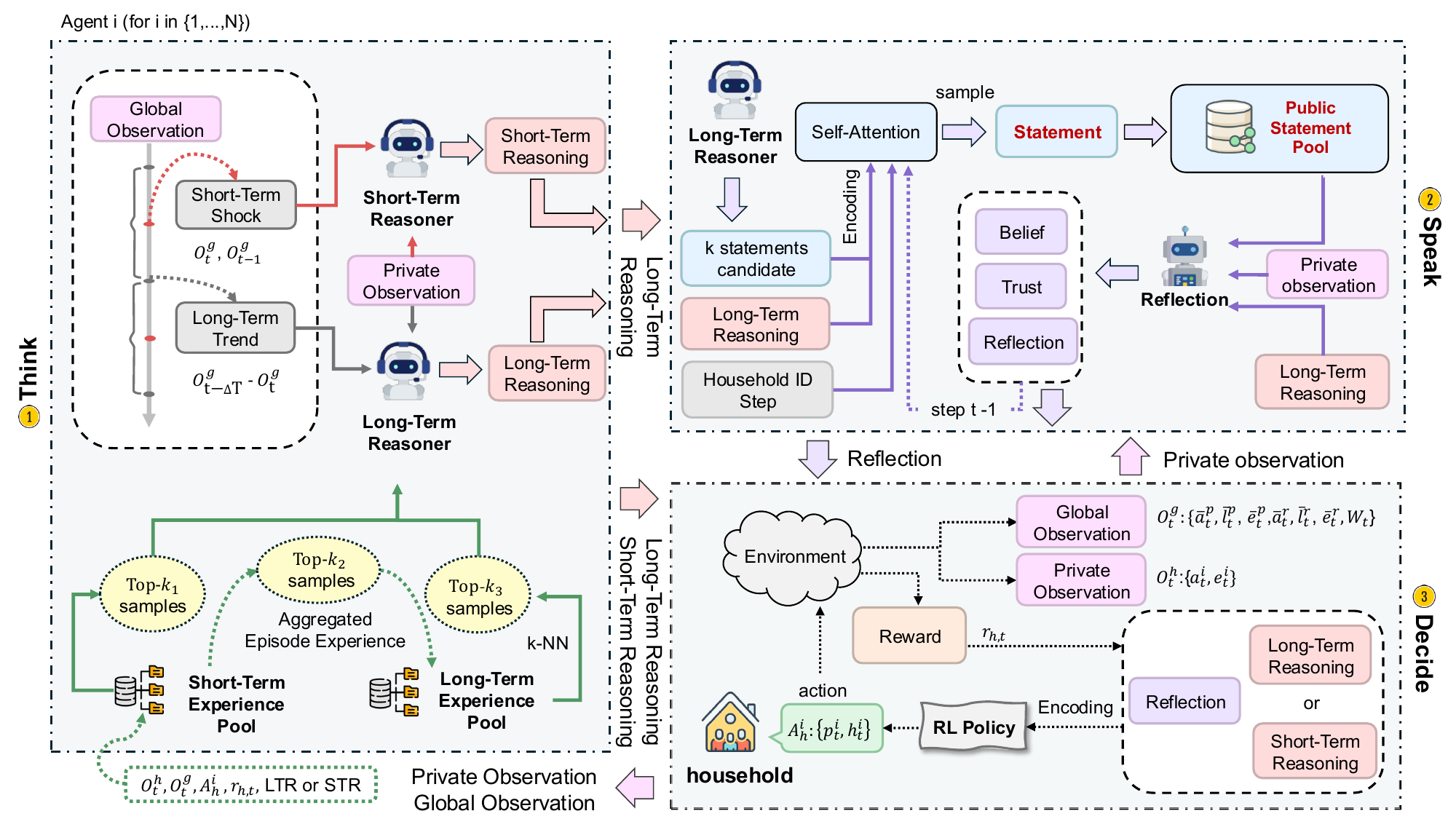} 
\caption{Workflow of LAMP: (a) Dual‑path Think module extracts long‑term trends and short‑term shocks into compact reasoning embeddings; (b) Speak module applies self‑attention to sample and broadcast a single message and performs a reflection step to update beliefs; (c) Decide module’s policy network concatenates numeric observations with language and reflection embeddings to select actions.}
\label{fig2}
\end{figure*}

This section first presents a mathematical formulation of the language-augmented multi-agent decision-making problem in economic environments (Section~\ref{problem}) and then details our proposed LAMP framework (Section~\ref{Framework}).

\subsection{Problem Formulation}\label{problem}
We formulate the economic decision‑making problem with language involvement. Building on the economic modeling in TaxAI~\cite{mi2023taxai}, we incorporate language by augmenting each household’s observation as
\[
  m_t^i = \mathcal{E}\bigl(\mathcal{L}(a_t^i, e_t^i, O_t^g))
\]
Here, $\mathcal{L}$ denotes a large language model producing a textual message from inputs, and $\mathcal{E}$ denotes an embedding model that maps this text into $\mathbb{R}^n$.
For inputs, all agents share a global observation \(O_t^g\).
The government observes
\(
O_t^g = \left\{ W_t,\, \bar{a}_t^{r,p},\, \bar{i}_t^{r,p},\, \bar{e}_t^{r,p} \right\},
\)
where $W_t$ denotes the wage, and the remaining terms are group-level averages of assets, income, and efficiency.
Each household $i$ observes the same \(O_t^g\) and, in addition, its private asset \(a_t^i\) and efficiency \(e_t^i\).

We then model the economic decision‑making problem as a partially observable Markov game
\(
\mathcal{M} = \bigl\langle N,\,S,\,O,\,A,\,P,\,R,\,\delta\bigr\rangle,
\)
where \(N = \{g,1,\dots,N_h\}\), \(\delta\in[0,1)\), and \(P\) is the transition kernel induced by
$ A = A^g \times A^{h,1} \times \cdots \times A^{h,N_h}.$
At each step, the government’s action is
$
A_t^g=\{\tau_t,\ \xi_t,\ \tau_{a,t},\ \xi_{a,t},\ r_t^{G}\},
$
where \(\tau_t\) and \(\xi_t\) parameterize the marginal income-tax schedule, \(\tau_{a,t}\) and \(\xi_{a,t}\) analogously parameterize the marginal asset-tax schedule, and \(r_t^{G}\) denotes the expenditure-to-output ratio.
Each household \(i\) selects a savings rate and labor supply \(h_t^i \in [0, h_{\max}]\):
$
A_t^{h,i} = \{p_t^i, h_t^i\}.
$

The government policy \(\pi_g\) and household policies \(\pi_i\) map their observations to actions.
The household’s objective is to maximize lifetime utility from consumption and leisure, with consumption increasing utility and labor hours reducing it:
\[
\max\;
\mathbb{E}_0 \sum_{t=0}^{T_N}
\beta^{t}\left(
  \frac{c_t^{1-\eta}}{1-\eta}
  - \frac{h_t^{1+\gamma}}{1+\gamma}
\right)
\]
\[
\text{s.t.} \quad (1+\tau_s)c_t + a_{t+1}
= i_t - T(i_t) + a_t - T^a(a_t)
\]
where \(c_t\) and \(h_t\) are consumption and labor, \(\beta\) is the discount factor, \(\eta\) is the relative risk aversion coefficient, and \(\gamma\) is the inverse Frisch elasticity.

The government’s objective is GDP growth; the government remains as in TaxAI, full details are provided in the Appendix~\ref{app:setup}.

\subsection{LAMP Framework}\label{Framework}

\begin{table}[t]
\centering
\small
\setlength{\tabcolsep}{4pt}
\begin{tabular}{ll}
\toprule
\textbf{Symbol} & \textbf{Description} \\
\midrule
\multicolumn{2}{l}{\textit{Economic Variables}} \\
$N_h$ & Number of households \\
$O_t^g$ & Government observation (wage, group averages) \\
$a_t^i,\, e_t^i$ & Asset, efficiency of household $i$ \\
$c_t,\, h_t$ & Consumption, labor \\
$\beta,\, \eta,\, \gamma$ & Discount, risk aversion, Frisch elasticity \\
$Y_t,\, G_t,\, B_t,\, T_t$ & GDP, spending, debt, tax \\
\midrule
\multicolumn{2}{l}{\textit{Framework Variables}} \\
$\mathcal{X}_t$ & Global indicators (Gini, welfare, GDP) \\
$\mathcal{L},\mathcal{E}$ & Language model, Embedding model \\
$\sigma,\, L_i$ & Shock threshold, long-term step size \\
$\mathcal{R}_t^{s},\, \mathcal{R}_{L_i}^{l}$ & Short-/long-term news \\
$\mathcal{H}^{s},\, \mathcal{H}^{l}$ & Short-/long-term experience \\
$\psi_t^{i},\, V_t$ & Reasoning, public statements \\
$m_t^i,\, x_t$ & Embedding, fused state \\
\bottomrule
\end{tabular}
\caption{Key symbols in the economic problem and LAMP.}
\label{tab:symbols_combined}
\end{table}

To address the above problem, we propose the LAMP framework (see Pseudocode~\ref{alg:lamp-core}), which comprises three modules:

\paragraph{Think}  
\textit{Think} translates global numerical signals into shared news, providing both short- and long-term economic interpretations to guide agents’ reasoning and dialogue.  
At fixed checkpoints $L_i$, it issues \textbf{long-term news} capturing structural trends. Whenever a key indicator \(
\mathcal{X}_t = \bigl(G_w,\,\mathcal{W},\,Y\bigr)
\) —wealth Gini $G_w$, social welfare $\mathcal{W}$, or per-capita GDP $Y$—changes by more than a threshold $\sigma$, it broadcasts a \textbf{short-term shock}. 
Then the news type is:  
\[
\text{type}(t) =
\begin{cases}
\text{long}, & t \in \{L_1, \dots, L_n\},\\[4pt]
\text{short}, & \displaystyle \max_j \bigl|\mathcal{X}_{j,t} - \mathcal{X}_{j,t-1}\bigr| > \sigma,\\[6pt]
\text{none}, & \text{otherwise}.
\end{cases}
\]  
This design ensures agents receive timely, context-rich updates—similar to how real-world economic actors rely on news outlets—rather than raw numerical data.  

A shared LLM-driven news service synthesizes appropriate texts $\mathcal{R}^{\text{short}}_t$ or $\mathcal{R}^{\text{long}}_{L_i}$ and disseminates them to all agents. Short-term news is generated as:  
\[
\mathcal{R}^{\text{short}}_t
= \mathcal{L}_{\text{S}}\!\left(O_t^g,\, O_{t-1}^g,\, \mathcal{R}^{\text{long}}_{L_k} \right), \quad  L_k < t < L_{k+1}
\]  
incorporating the current and previous global observations, as well as the most recent long-term news. Long-term news is generated over a two-step observation window:  
\[
\mathcal{R}^{\text{long}}_{L_i}
= \mathcal{L}_{\text{L}}\!\left(O_{L_i-1:L_i}^g\right), \quad i = 1,2,\dots,n
\]  

Upon receiving short-term news, each agent infers its economic status $\kappa_t^{\,i} \in \{0,1,2\}$ (good / neutral / poor) and produces a private reasoning $\psi_t^{\,i}$. Long-term news additionally triggers the Experience Pool and Speak module for deeper reasoning.  
After each short-term reasoning phase, agent $i$ ranks candidate reasoning trajectories by reward and stores its top $k_1$ reasoning trajectories into a \emph{short-term} buffer:  
\[
\mathcal{H}_{t,i}^{\text{short}} = \mathrm{Top}_{k_1}(\mathcal{T}_i)
\]  
At each long-term checkpoint $k$, the system collects the top $k_2$ trajectories (by reward) across all agents and appends them to the \emph{long-term} FAISS index:  
\[
\mathcal{H}_{k}^{\text{long}} = \mathcal{H}_{k-1}^{\text{long}} \cup \mathrm{Top}_{k_2}\bigl(\bigcup_{i=1}^{N_h}\mathcal{T}_i\bigr)
\]  

Before the next long-term reasoning step, agent \(i\) retrieves the \(k_3\) nearest neighbors from \(\mathcal{H}_{k}^{\text{long}}\) using FAISS, where similarity is computed against a query embedding derived from its current observation \(O_t^{\,h,i}\), and merges them with its current \(\mathcal{H}_{t,i}^{\text{short}}\). 
This combined set of past high-reward insights is then used as contextual prompts for the LLM:  
\[
\mathcal{H}_{k,t}^{i} = \mathrm{kNN}_{k_3}(\mathcal{H}_k^{\text{long}}) \cup \mathcal{H}_{t,i}^{\text{short}}
\]  
allowing the agent to remember and reuse successful strategies in similar future scenarios.

\begin{figure*}[t]
\centering
\includegraphics[width=0.95\textwidth]{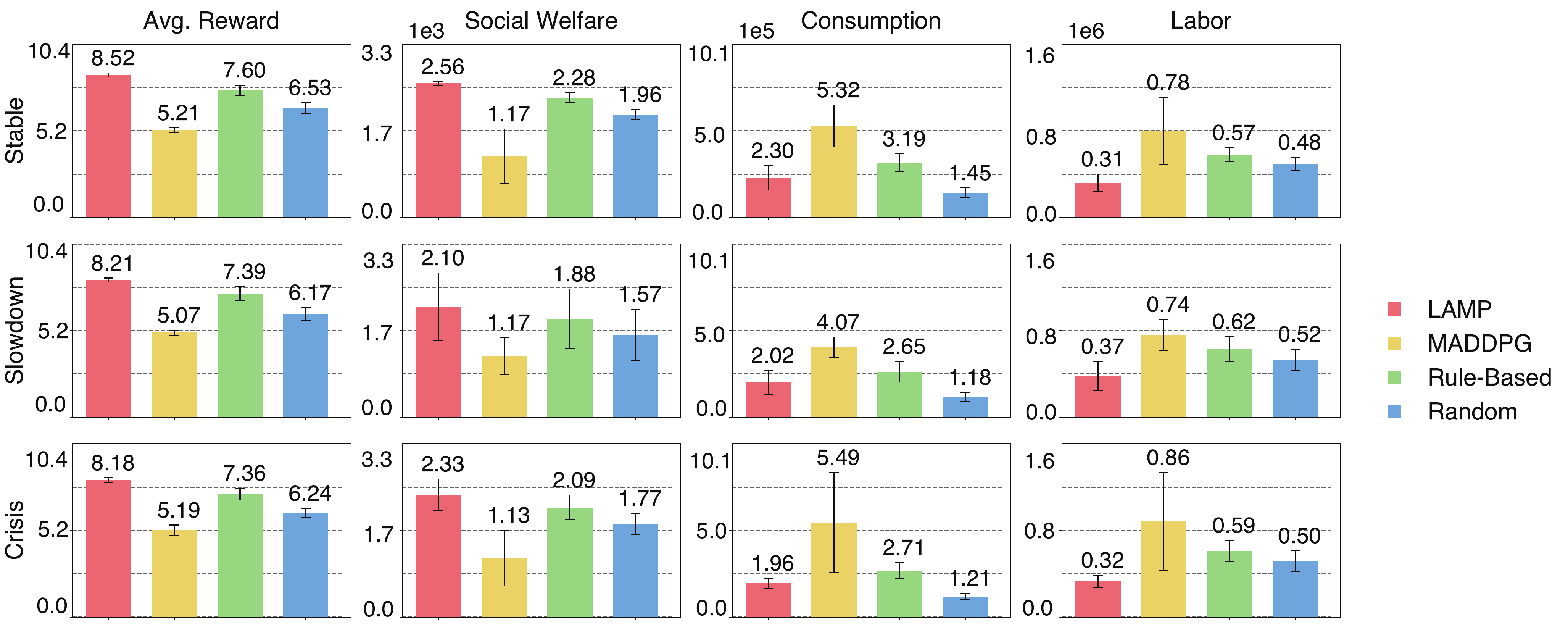} 
\caption{Across three economic environments, LAMP outperforms non-language baselines (Random, rule-based, MADDPG) with higher social welfare and reward, and similar labor usage.}
\label{fig3}
\end{figure*}

\begin{figure*}[t]
\centering
\includegraphics[width=1\textwidth]{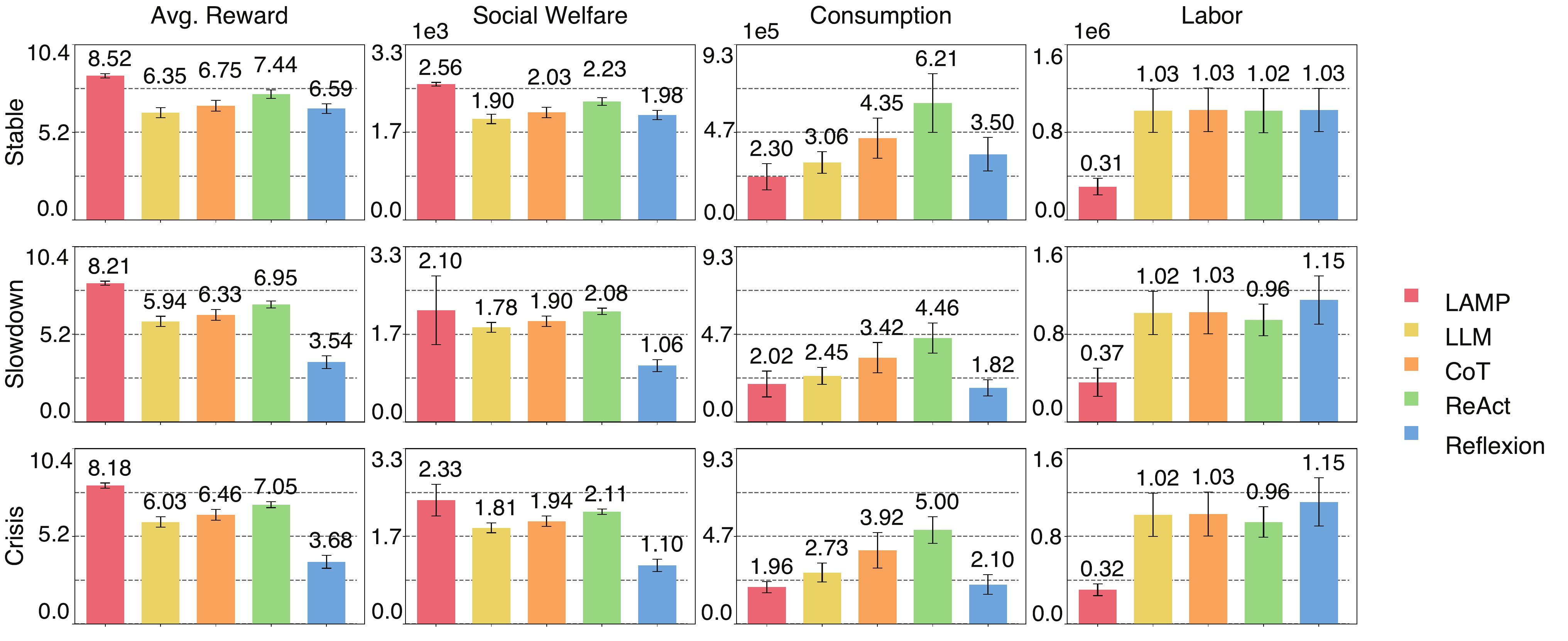} 
\caption{LAMP vs. other language-based agents (Only-LLM, CoT, ReAct, Reflexion) on the same metrics across the three economic environments. LAMP outperforms all these LLM-driven baselines, obtaining higher social welfare and reward.}
\label{fig4}
\end{figure*}

\paragraph{Speak}  
Building on the news from \textit{Think} and each agent’s private reasoning, \textit{Speak} produces a concise strategic statement per agent, broadcasts it to peers, and returns language-based peer assessments for the next reasoning step.

Inspired by ~\cite{xu2023language}, the LLM generates three candidate statements for agent \(i\); a self-attention selector \(\mathcal{S}\) scores them to form a distribution \(p_t^{\,i,\cdot}\), from which one statement is sampled and broadcast to all agents. Let \(V_t\) denote the multiset of broadcast statements. After broadcasting and receiving messages \(V_t\), each agent \(i\) uses a Reflection Module \(\mathcal{L}_{reflect}\) to interpret the content. This produces an assessment of each peer \(j\), including an estimated wealth tier (\(w_t^{\,i\!\to j} \in \{\text{low, mid, high}\}\)) and a numeric belief confidence \(\tau_t^{\,i\!\to j} \in [0,10]\). The evaluator also generates a brief self-reflection \(\alpha_t^{\,i}\) summarizing agent \(i\)’s own situation:  
\[
\bigl(w_{t}^{\,i\!\rightarrow j},\, \tau_{t}^{\,i\!\rightarrow j},\, \alpha_t^{\,i}\bigr)
= \mathcal{L}_{\mathrm{reflect}}\!\bigl(O_t^{\,h,i}, V_t, \psi_t^{\,i}\bigr)
\]  

These peer assessments are fed back to \(\mathcal{S}\) and the LLM policy to guide the next round of reasoning and candidate selection, closing a loop that links language reasoning, dialogue, and adaptive coordination.  

\paragraph{Decide}  
Consuming language embeddings from \textit{Think} and \textit{Speak} together with numeric observations, \textit{Decide} compresses language vectors and maps the enriched state to actions under centralized training with decentralized execution (CTDE).
All texts (private reasoning and reflection) are encoded by a text encoder \(\mathcal{E}_{\text{text}}\), pooled into a fixed-length vector \(h_t^{\,i}\), and passed through a small projection \(P:\mathbb{R}^{D}\!\to\!\mathbb{R}^{d}\) for dimensionality reduction and feature alignment:  
\[
\tilde{m}_t^i
=\frac{P(h_t^{\,i})}{\|P(h_t^{\,i})\|_2}\in\mathbb{R}^{d}.
\]
Unless otherwise noted, gradients do not flow into \(\mathcal{E}_{\text{text}}\) (the encoder is frozen for stability) and only \(P(\cdot)\) is updated during RL. 
At time \(t\), the observation is concatenated with household language embeddings to form  $x_t = \bigl(O_t^g,\; o_{t}^{\,1:\!N_h},\; m_t^{\,1:\!N_h}\bigr),$
which, together with the joint action \(a_t\), is stored in the replay buffer \(\mathcal{D}\).  
We adopt a standard MADDPG framework~\cite{lowe2017multi}, where a centralized critic minimizes Bellman error, and decentralized actors update their policies by maximizing the expected \(Q\)-value via deterministic policy gradients.  
Full optimization details and loss formulations are provided in the Appendix~\ref{app:setup}.

\begin{table*}[t]
  \centering
  \small
  \begin{tabular}{llcccc}
    \toprule
    \textbf{Category} & \textbf{Algorithms} & \textbf{Avg. Reward} ( $\uparrow$ ) & \textbf{Social Welfare} ( $\uparrow$ ) & \textbf{Consumption} ( - ) & \textbf{Labor} ( - ) \\
    \midrule
    \textbf{Ours} & \textbf{LAMP}
      & \textbf{8.52 $\pm$ 0.13}
      & \textbf{2.56e+03 $\pm$ 3.77e+01}
      & 2.30e+05 $\pm$ 7.52e+04
      & 3.13e+05 $\pm$ 8.46e+04 \\
    \midrule
    \textbf{Conventional} & \textbf{MADDPG}
      & 5.21 $\pm$ 0.16
      & 1.17e+03 $\pm$ 5.51e+02
      & 5.32e+05 $\pm$ 1.31e+05
      & 7.82e+05 $\pm$ 3.20e+05 \\
    ~ & \textbf{Rule-Based}
      & 7.60 $\pm$ 0.33
      & 2.28e+03 $\pm$ 9.99e+01
      & 3.19e+05 $\pm$ 5.46e+04
      & 5.68e+05 $\pm$ 6.73e+04 \\
    ~ & \textbf{Random}
      & 6.53 $\pm$ 0.35
      & 1.96e+03 $\pm$ 1.06e+02
      & 1.45e+05 $\pm$ 3.10e+04
      & 4.84e+05 $\pm$ 6.41e+04 \\
    \midrule
    \textbf{LLM-based} & \textbf{LLM-Only}
      & 6.35 $\pm$ 0.32
      & 1.90e+03 $\pm$ 9.56e+01
      & 3.06e+05 $\pm$ 6.14e+04
      & 1.03e+06 $\pm$ 2.18e+05 \\
    ~ & \textbf{CoT}
      & 6.75 $\pm$ 0.34
      & 2.03e+03 $\pm$ 1.03e+02
      & 4.35e+05 $\pm$ 1.14e+05
      & 1.03e+06 $\pm$ 2.19e+05 \\
    ~ & \textbf{ReAct}
      & 7.44 $\pm$ 0.26
      & 2.23e+03 $\pm$ 7.92e+01
      & 6.21e+05 $\pm$ 1.66e+05
      & 1.02e+06 $\pm$ 2.21e+05 \\
    ~ & \textbf{Reflexion}
      & 6.59 $\pm$ 0.31
      & 1.98e+03 $\pm$ 9.16e+01
      & 3.50e+05 $\pm$ 9.51e+04
      & 1.03e+06 $\pm$ 2.16e+05 \\
    \bottomrule
  \end{tabular}
  \caption{Comparison of LAMP with conventional and LLM-based baselines in the real-data–calibrated environment (S1: Economic Stability). Results for S2 and S3 appear in the Appendix~\ref{app:setup}.
Values are mean $\pm$ SD. 
Notation: $(\uparrow)$ higher is better; $(\text{--})$ non-monotonic. 
Consumption and Labor jointly shape household utility with non-monotonic effects.
}
  \label{tab:baseline_methods_compact}
\end{table*}

\begin{table*}[t]
  \centering
  \small
  \begin{tabular}{lccccc}
    \toprule
    \textbf{Ablation Setting} & \textbf{Avg. Reward} ( $\uparrow$ ) & \textbf{Social Welfare} ( $\uparrow$ ) & \textbf{Consumption} ( - ) & \textbf{Labor} ( - ) & \textbf{Years} ( $\uparrow$ ) \\
    \midrule
    \textbf{LAMP (Ours)}
      & \textbf{8.52} & \textbf{2.56e+03} & 2.30e+05 & 3.13e+05 & \textbf{3.00e+02} \\
    \midrule
    \textbf{\,\,w/o Speak}
      & 8.42 ({$-1\%$})
      & 2.53e+03 ({$-1\%$})
      & 3.24e+05 ({$+41\%$})
      & 5.36e+05 ({$+71\%$})
      & \textbf{3.00e+02} ({$+0\%$}) \\
    \textbf{\,\,w/o Experience Pool}
      & 8.45 ({$-1\%$})
      & 1.25e+03 ({$-51\%$})
      & 5.12e+05 ({$+122\%$})
      & 4.50e+05 ({$+44\%$})
      & 1.50e+02 ({$-50\%$}) \\
    \textbf{\,\,w/o Long-Term}
      & 5.31 ({$-38\%$})
      & 1.15e+03 ({$-55\%$})
      & 2.27e+05 ({$-2\%$})
      & 4.10e+05 ({$+31\%$})
      & 2.19e+02 ({$-27\%$}) \\
    \textbf{\,\,w/o Short-Term}
      & 8.18 ({$-4\%$})
      & 1.67e+03 ({$-35\%$})
      & 3.51e+05 ({$+53\%$})
      & 5.25e+05 ({$+68\%$})
      & 2.08e+02 ({$-30\%$}) \\
    \textbf{\,\,w/o Timing Scheduler}
      & \textbf{8.52} ({$-0\%$})
      & 1.19e+03 ({$-53\%$})
      & 3.48e+05 ({$+51\%$})
      & 5.70e+05 ({$+82\%$})
      & 1.41e+02 ({$-53\%$}) \\
    \bottomrule
  \end{tabular}
  \label{tab:ablation_metrics_compact}
  \caption{Ablation under the baseline economy. Percentages denote change vs. LAMP (Ours). Notation: $(\uparrow)$ higher is better; $(\text{--})$ non-monotonic. Consumption and Labor jointly shape household utility with non-monotonic effects.}
\end{table*}

\section{Experiments}

Our experiments address two key questions:  
\begin{enumerate}[leftmargin=*,itemsep=0pt,topsep=0pt]
\item \textbf{How effective is LAMP?} (\S~\ref{sec:compare}) We compare LAMP with non-language and LLM-based baselines across 3 economic scenarios to evaluate its performance.  
\item \textbf{What drives LAMP’s gains?} (\S~\ref{sec:ablation}) We remove core modules of LAMP to assess their contribution to performance and stability.
\end{enumerate}

\subsection{Experimental Setup}
\label{sec:setup}

\paragraph{Environment}
All experiments are conducted in \textbf{TaxAI}~\cite{mi2023taxai}, a dynamic economic simulator. It models complex economic interactions between heterogeneous households and a government, and is calibrated with real-world data—making it a realistic and challenging testbed for economic decision-making.

\paragraph{Evaluation Metrics}
We evaluate LAMP and baselines with five metrics: (1) \textbf{Average Household Reward} — mean reward per step across households; (2) \textbf{Social Welfare}: sum of utilities across all households over the horizon; (3) \textbf{Total Consumption}: aggregate consumption of households; (4) \textbf{Total Labor}: aggregate labor supply in an economy; and (5) \textbf{Years}: number of simulated years before collapse (max 300, higher indicates greater stability). \textit{Total Consumption} and \textit{Total Labor} do not directly measure policy performance, but help analyze policy preferences.

\paragraph{Baselines.}
We benchmark LAMP against two baseline categories with identical training budgets and horizons. All LLM-based baselines use the same backbone (\textbf{Qwen2.5-72B-Instruct-INT4}) and prompts. We compare different language models in the Appendix~\ref{app:setup}.

\textbf{(1) Conventional Baselines:}
\textbf{Random}: Agents select actions uniformly at random.
\textbf{Rule-Based}: Economic method based on the utility–production model (details in extended version).
\textbf{MADDPG}: Multi-Agent Deep Deterministic Policy Gradient~\cite{lowe2017multi}. We also compare different MARL algorithms in the Appendix~\ref{app:setup}.

\textbf{(2) LLM-based Baselines:}
 \textbf{Only-LLM}: Directly query an LLM to generate actions.
 \textbf{CoT / ReAct / Reflexion}: LLM reasoning methods using CoT~\cite{wei2022chain}, ReAct~\cite{yao2023react}, or Reflexion~\cite{shinn2023reflexion}.

\subsection{How effective is LAMP?}
\label{sec:compare}

We evaluate LAMP and baselines under three settings:
\begin{itemize}[leftmargin=1.2em, itemsep=0pt, topsep=0pt]
\item \textbf{Economic Stability (S1):} Matches training conditions, representing a stable macroeconomic scenario.
\item \textbf{Economic Slowdown (S2):} Introduces a moderate shift, simulating reduced growth and mild market stress.
\item \textbf{Crisis Shock (S3):} Applies a large, coupled shift, modeling severe economic shocks for robustness evaluation.
\end{itemize}
Detailed setup is provided in the Appendix~\ref{app:setup}.

\paragraph{Quantifying Gains over LLM-based Baselines.}
LAMP also outperforms language-integrated baselines, demonstrating the advantage of combining MARL with language-guided policy optimization.
In \textbf{S1}, using the same backbone and prompt budget, LAMP surpasses the strongest language baseline (ReAct) with \textbf{+14.8\%} higher welfare and \textbf{+14.5\%} higher reward, while reducing consumption and labor.
Under distribution shifts, the advantage remains: in \textbf{S2} and \textbf{S3}, welfare gains are \textbf{+1.0\%} and \textbf{+10.4\%}, reward gains are \textbf{+18.1\%} and \textbf{+16.0\%}, with corresponding reductions in consumption and labor.
These results confirm that LAMP’s language-guided coordination improves both stability and efficiency, even in stressed economic conditions.

\paragraph{Isolating Language Effects.}
LAMP consistently outperforms non-language baselines, demonstrating the benefit of language integration in economic decision-making.
In \textbf{S1}, LAMP achieves the highest \emph{Social Welfare} and \emph{Average Household Reward}. Compared to the strongest non-language baseline (Rule-Based), welfare improves by \textbf{+12.3\%} and reward by \textbf{+12.1\%}; relative to numeric MARL (MADDPG), gains reach \textbf{+118.8\%} and \textbf{+63.5\%}, respectively.
Efficiency gains are evident from lower \emph{Consumption} and \emph{Labor}. Versus Rule-Based, LAMP uses –27.9\% consumption and –44.9\% labor (vs.\ MADDPG: –56.8\% and –60.0\%), suggesting that higher welfare stems from efficiency rather than brute-force spending or overwork. Under \textbf{S2} and \textbf{S3}, LAMP consistently outperforms the baselines.

\paragraph{Analysis and Insights.}
We share \textbf{interesting findings} from experiments, supported by LLM outputs:

(1) Economic decision‑making involves many interdependent variables that change frequently, with causal links often unclear. Purely data‑driven MARL starts from scratch, fitting policies without explicit understanding of these variables, making optimal policy search slow and uncertain.

(2) LAMP addresses this by using LLM reasoning at each step to extract concise, high‑value insights, which are then passed to the MARL component. These structured signals—hard for pure data‑driven methods to obtain—are readily produced by LLMs.
Representative examples illustrate the LLM’s clear interpretation of economic variables and targeted reasoning that enhance decision‑making. More examples are shown in the Appendix~\ref{app:text_examples}.
\subsection{What drives LAMP’s gains?}
\label{sec:ablation}

\paragraph{Speak Module: Strategy Communication \& Opponent Modeling.}
The \textit{Speak} module enables agents to exchange strategic messages and infer others’ states, providing the coordination essential for high performance. Removing it causes a 1.2\% welfare drop alongside sharp increases in labor and consumption. This indicates that, without strategic communication, agents compensate through brute‑force effort. With Speak enabled, comparable or higher welfare is achieved with far less input. Representative outputs show the mechanism: after detecting widening inequality and low wages, the LLM revises beliefs toward demand fragility and restraint, then recommends disciplined actions such as moderating labor expansion and investing in human capital, thereby reducing overshooting and volatility.

\paragraph{Experience Pool: Enhancing Stability and Efficiency.}
The experience pool substantially improves efficiency and stability. Removing it \textbf{cuts social welfare by 50.9}\% and average household reward by 0.8\%, while labor rises 43.6\% and consumption surges 122.4\%. The unexpected jump in consumption suggests that, without stored successful trajectories, agents overshoot spending and output, oscillating in search of workable strategies. Stability also deteriorates, with \textbf{50.2\% fewer simulated years} sustained before failure. Beyond performance, the pool improves interpretability by preserving reasoning traces as an auditable knowledge base explaining why certain strategies are followed.

\paragraph{Reasoning Paths: Trend Tracking \& Shock Response.}
\textit{Long‑term reasoning} is essential for capturing structural trends. Removing it \textbf{drops average household reward by 37.7\%} and reduces stable years from 300 to 219. Without long‑term reasoning, agents become myopic, reacting only to immediate stimuli and producing unstable policies.

\textit{Short‑term reasoning} supports rapid adjustment to shocks. Disabling it has a moderate effect on final returns (\textbf{–3.99\% reward}) but significantly harms efficiency: labor rises 67.7\%, consumption 52.7\%, and stable years fall from 300 to 208.

\textit{Trigger timing} is critical. Random triggers keep welfare similar but collapse efficiency: labor increases 81.9\%, consumption 51.2\%, and stable years decrease from \textbf{300} to \textbf{141}. This shows aligning reasoning with actual needs reduces turbulence and sustains consistent performance.

\textbf{We observe an adaptive policy shift in LLM outputs}: upon detecting rising inequality—top 10\% volatility widening and bottom 50\% declining—the LLM revised its earlier “work more” stance. It recommended slightly reducing work hours, increasing savings, delaying non‑essential spending, and investing in skills for long‑term stability, while publicly supporting progressive taxation and minimum wages.

\begin{tcolorbox}[title=\textbf{Representative LLM Reasoning and Experience}, breakable]
\small
\begin{description}[leftmargin=0pt, labelsep=0.6em, style=nextline, font=\ttfamily]

  \item[Short-term]
  \textbf{Reasoning:} ``... The family's personal productivity (0.7741) and wealth (0.0957) place them in a vulnerable position. Given the volatility and risk of instability, the economic status is rated as `Bad'.''\\
  \textbf{Economic status:} \texttt{0}

  \item[Long-term]
  \textbf{Statement:} ``We should advocate for policies that promote fair wage growth and equitable wealth distribution to stabilize the broader economic environment and ...''\\
  \textbf{Reasoning:} ``The family should avoid overwork and instead focus on savings, education, and...''\\
  \textbf{Reflection:} ``...highlight the importance of balancing increased labor time... Investing in education and advocating for fairness improves resilience and security.''\\
  \textbf{Economic status:} \texttt{1}\\
  \textbf{Belief:} \texttt{[0, 1, 0, 0, 0, 1, 1, 1, 0, 2]}\\
  \textbf{Trust:} \texttt{[8, 9, 9, 8, 9, 8, 9, 9, 8, 10]}

\item[Experience]
     ID=Household1,
     Reward=0.95,
     Personal productivity(e): 1.846,
     Personal wealth: 0.196, 
     savings ratio:-0.947,
     working time ratio:-0.963,
     Reasoning: ``...''

\end{description}
\end{tcolorbox}

\section{Conclusion}
This paper introduced the \textbf{Language‑Augmented Multi‑Agent Policy (LAMP)} framework, offering a new approach to complex economic decision‑making. LAMP leverages LLM reasoning and reflection over language signals—such as peer dialogue and public news—alongside numerical data to inform optimal policies.
The framework follows a \textit{Think–Speak–Decide} pipeline: agents extract short‑term shocks and long‑term trends, communicate strategic insights, and execute language‑informed policies. \textbf{Experiments demonstrate LAMP’s strong performance and reveal interesting insights}: LLM reasoning and reflection dynamically distill key information from numerous, volatile economic variables, enabling agents to make efficient decisions.

This contrasts with fully data-driven methods that search for optimal solutions from scratch—a process particularly challenging in economics. We hope this work offers novel methods and insights for AI in economic decision-making.

\section*{Acknowledgments}
We sincerely thank Prof. Hao Huang from the University of International Business and Economics for his valuable guidance and insightful suggestions during the early stage of this work. 

This work was supported in part by the National Natural Science Foundation of China under the Original Exploration Program (Grant No. 72450002).

\bibliography{aaai2026}

@book{varian1992microeconomic,
  title={Microeconomic analysis},
  author={Varian, Hal R and Varian, Hal R},
  volume={3},
  year={1992},
  publisher={Norton New York}
}

@article{zheng2022ai,
  title={The AI Economist: Taxation policy design via two-level deep multiagent reinforcement learning},
  author={Zheng, Stephan and Trott, Alexander and Srinivasa, Sunil and Parkes, David C and Socher, Richard},
  journal={Science Advances},
  volume={8},
  number={18},
  pages={eabk2607},
  year={2022},
  publisher={American Association for the Advancement of Science}
}

@inproceedings{mi2023taxai,
  title={TaxAI: A Dynamic Economic Simulator and Benchmark for Multi-agent Reinforcement Learning},
  author={Mi, Qirui and Xia, Siyu and Song, Yan and Zhang, Haifeng and Zhu, Shenghao and Wang, Jun},
  booktitle={Proceedings of the 23rd International Conference on Autonomous Agents and Multiagent Systems},
  pages={1390--1399},
  year={2024}
}

@inproceedings{li2024econagent,
  title={EconAgent: Large Language Model-Empowered Agents for Simulating Macroeconomic Activities},
  author={Li, Nian and Gao, Chen and Li, Mingyu and Li, Yong and Liao, Qingmin},
  booktitle={Proceedings of the 62nd Annual Meeting of the Association for Computational Linguistics (Volume 1: Long Papers)},
  pages={15523--15536},
  year={2024}
}

@techreport{horton2023large,
  title={Large language models as simulated economic agents: What can we learn from homo silicus?},
  author={Horton, John J},
  year={2023},
  institution={National Bureau of Economic Research}
}

@inproceedings{park2023generative,
  title={Generative agents: Interactive simulacra of human behavior},
  author={Park, Joon Sung and O'Brien, Joseph and Cai, Carrie Jun and Morris, Meredith Ringel and Liang, Percy and Bernstein, Michael S},
  booktitle={Proceedings of the 36th Annual ACM Symposium on User Interface Software and Technology},
  pages={1--22},
  year={2023}
}

@article{hao2025multi,
  title={A Multi-LLM-Agent-Based Framework for Economic and Public Policy Analysis},
  author={Hao, Yuzhi and Xie, Danyang},
  journal={arXiv preprint arXiv:2502.16879},
  year={2025}
}

@article{douglas2024consumption,
  title={Consumption and Savings with Large Language Model Agents},
  author={Douglas, Michael R and Verstyuk, Sergiy},
  journal={Available at SSRN 4909749},
  year={2024}
}

@article{xiao2024tradingagents,
  title={TradingAgents: Multi-Agents LLM Financial Trading Framework},
  author={Xiao, Yijia and Sun, Edward and Luo, Di and Wang, Wei},
  journal={arXiv preprint arXiv:2412.20138},
  year={2024}
}

@article{yu2024fincon,
  title={Fincon: A synthesized llm multi-agent system with conceptual verbal reinforcement for enhanced financial decision making},
  author={Yu, Yangyang and Yao, Zhiyuan and Li, Haohang and Deng, Zhiyang and Jiang, Yuechen and Cao, Yupeng and Chen, Zhi and Suchow, Jordan and Cui, Zhenyu and Liu, Rong and others},
  journal={Advances in Neural Information Processing Systems},
  volume={37},
  pages={137010--137045},
  year={2024}
}

@book{atashbar2023ai,
  title={AI and macroeconomic modeling: Deep reinforcement learning in an RBC model},
  author={Atashbar, Tohid and Shi, Rui Aruhan},
  year={2023},
  publisher={International Monetary Fund}
}

@misc{chenDeepReinforcementLearning2023,
  title = {Deep {{Reinforcement Learning}} in a {{Monetary Model}}},
  author = {Chen, Mingli and Joseph, Andreas and Kumhof, Michael and Pan, Xinlei and Zhou, Xuan},
  year = {2023},
  month = jan,
  number = {arXiv:2104.09368},
  eprint = {2104.09368},
  primaryclass = {econ, q-fin, stat},
  publisher = {{arXiv}},
  doi = {10.48550/arXiv.2104.09368},
  urldate = {2023-05-10},
  archiveprefix = {arxiv}
}

@misc{danassisAIdrivenPricesExternalities2023,
  title = {{{AI-driven Prices}} for {{Externalities}} and {{Sustainability}} in {{Production Markets}}},
  author = {Danassis, Panayiotis and {Filos-Ratsikas}, Aris and Chen, Haipeng and Tambe, Milind and Faltings, Boi},
  year = {2023},
  month = jan,
  number = {arXiv:2106.06060},
  eprint = {2106.06060},
  primaryclass = {cs},
  publisher = {{arXiv}},
  urldate = {2023-03-16},
  archiveprefix = {arxiv}
}

@inproceedings{mi2025learning,
  title={Learning Macroeconomic Policies through Dynamic Stackelberg Mean-Field Games},
  author={Mi, Qirui and Zhao, Zhiyu and Ma, Chengdong and Xia, Siyu and Song, Yan and Yang, Mengyue and Wang, Jun and Zhang, Haifeng},
  booktitle={28th European Conference on Artificial Intelligence (ECAI) 2025},
  year={2025}
}

@misc{johansonEmergentBarteringBehaviour2022,
  title = {Emergent {{Bartering Behaviour}} in {{Multi-Agent Reinforcement Learning}}},
  author = {Johanson, Michael Bradley and Hughes, Edward and Timbers, Finbarr and Leibo, Joel Z.},
  year = {2022},
  month = may,
  number = {arXiv:2205.06760},
  eprint = {2205.06760},
  primaryclass = {cs},
  publisher = {{arXiv}},
  doi = {10.48550/arXiv.2205.06760},
  urldate = {2023-05-11},
  archiveprefix = {arxiv}
}

@misc{kurikshaEconomyNeuralNetworks2021,
  title = {An {{Economy}} of {{Neural Networks}}: {{Learning}} from {{Heterogeneous Experiences}}},
  shorttitle = {An {{Economy}} of {{Neural Networks}}},
  author = {Kuriksha, Artem},
  year = {2021},
  month = oct,
  number = {arXiv:2110.11582},
  eprint = {2110.11582},
  primaryclass = {cs, econ, q-fin},
  publisher = {{arXiv}},
  doi = {10.48550/arXiv.2110.11582},
  urldate = {2023-03-22},
  archiveprefix = {arxiv}
}

@misc{rui2022learningzeromakeconsumptionsaving,
      title={Learning from zero: how to make consumption-saving decisions in a stochastic environment with an AI algorithm}, 
      author={Rui (Aruhan) Shi},
      year={2022},
      eprint={2105.10099},
      archivePrefix={arXiv},
      primaryClass={econ.TH},
      url={https://arxiv.org/abs/2105.10099}, 
}

@misc{rui2022aiagenthitmoving,
      title={Can an AI agent hit a moving target?}, 
      author={Rui (Aruhan) Shi},
      year={2022},
      eprint={2110.02474},
      archivePrefix={arXiv},
      primaryClass={econ.TH},
      url={https://arxiv.org/abs/2110.02474}, 
}

@article{sch2021intelligence,
  title={Intelligence in the Economy: Emergent Behaviour in International Trade Modelling with Reinforcement Learning},
  author={Sch, Abraham Ayooluwa Odukoya},
  year={2021}
}

@article{charpentier2023reinforcement,
  title={Reinforcement learning in economics and finance},
  author={Charpentier, Arthur and Elie, Romuald and Remlinger, Carl},
  journal={Computational Economics},
  volume={62},
  number={1},
  pages={425--462},
  year={2023},
  publisher={Springer}
}

@article{luketina2019survey,
  title={A survey of reinforcement learning informed by natural language},
  author={Luketina, Jelena and Nardelli, Nantas and Farquhar, Gregory and Foerster, Jakob and Andreas, Jacob and Grefenstette, Edward and Whiteson, Shimon and Rockt{\"a}schel, Tim},
  journal={arXiv preprint arXiv:1906.03926},
  year={2019}
}

@article{zhu2024survey,
  title={A survey of multi-agent deep reinforcement learning with communication},
  author={Zhu, Changxi and Dastani, Mehdi and Wang, Shihan},
  journal={Autonomous Agents and Multi-Agent Systems},
  volume={38},
  number={1},
  pages={4},
  year={2024},
  publisher={Springer}
}

@misc{ozhamaratliDeepReinforcementLearning2022a,
  title = {Deep {{Reinforcement Learning}} for {{Optimal Investment}} and {{Saving Strategy Selection}} in {{Heterogeneous Profiles}}: {{Intelligent Agents}} Working towards Retirement},
  shorttitle = {Deep {{Reinforcement Learning}} for {{Optimal Investment}} and {{Saving Strategy Selection}} in {{Heterogeneous Profiles}}},
  author = {Ozhamaratli, Fatih and Barucca, Paolo},
  year = {2022},
  month = jun,
  number = {arXiv:2206.05835},
  eprint = {2206.05835},
  primaryclass = {cs, econ, q-fin},
  publisher = {{arXiv}},
  urldate = {2023-05-10},
  archiveprefix = {arxiv}
}

@article{tversky1974judgment,
  title={Judgment under Uncertainty: Heuristics and Biases: Biases in judgments reveal some heuristics of thinking under uncertainty.},
  author={Tversky, Amos and Kahneman, Daniel},
  journal={science},
  volume={185},
  number={4157},
  pages={1124--1131},
  year={1974},
  publisher={American association for the advancement of science}
}

@inproceedings{zhao2024mean,
  title={Mean Field Correlated Imitation Learning},
  author={Zhao, Zhiyu and Ma, Chengdong and Mi, Qirui and Yang, Ning and Yan, Xue and Yang, Mengyue and Zhang, Haifeng and Wang, Jun and Yang, Yaodong},
  booktitle={Proceedings of the 24th International Conference on Autonomous Agents and Multiagent Systems},
  pages={2364--2372},
  year={2025}
}

@inproceedings{mi2025econgym,
  title={EconGym: A Scalable AI Testbed with Diverse Economic Tasks},
  author={Mi, Qirui and Yang, Qipeng and Fan, Zijun and Fan, Wentian and Ma, Heyang and Ma, Chengdong and Xia, Siyu and An, Bo and Wang, Jun and Zhang, Haifeng},
  booktitle = {Advances in Neural Information Processing Systems (NeurIPS) 2025},
  year={2025}
}

@inproceedings{mi2025mf,
  title={MF-LLM: Simulating Population Decision Dynamics via a Mean-Field Large Language Model Framework},
  author={Mi, Qirui and Yang, Mengyue and Yu, Xiangning and Zhao, Zhiyu and Deng, Cheng and An, Bo and Zhang, Haifeng and Chen, Xu and Wang, Jun},
  booktitle = {Advances in Neural Information Processing Systems (NeurIPS) 2025},
  year={2025}
}

@misc{
slumbers2024leveraging,
title={Leveraging Large Language Models for Optimised Coordination in Textual Multi-Agent Reinforcement Learning},
author={Oliver Slumbers and David Henry Mguni and Kun Shao and Jun Wang},
year={2024},
url={https://openreview.net/forum?id=1PPjf4wife}
}

@ARTICLE{11027664,
  author={Zhu, Guobin and Zhou, Rui and Ji, Wenkang and Zhao, Shiyu},
  journal={IEEE Robotics and Automation Letters}, 
  title={LAMARL: LLM-Aided Multi-Agent Reinforcement Learning for Cooperative Policy Generation}, 
  year={2025},
  volume={10},
  number={7},
  pages={7476-7483},
  doi={10.1109/LRA.2025.3577527}}

@inproceedings{park2025maporl,
  title={Maporl: Multi-agent post-co-training for collaborative large language models with reinforcement learning},
  author={Park, Chanwoo and Han, Seungju and Guo, Xingzhi and Ozdaglar, Asuman E and Zhang, Kaiqing and Kim, Joo-Kyung},
  booktitle={Proceedings of the 63rd Annual Meeting of the Association for Computational Linguistics (Volume 1: Long Papers)},
  pages={30215--30248},
  year={2025}
}

@inproceedings{ma2024cory,
  title     = {Coevolving with the Other You: Fine-Tuning LLM with Sequential Cooperative Multi-Agent Reinforcement Learning},
  author    = {Ma, Hao and Hu, Tianyi and Pu, Zhiqiang and Liu, Boyin and Ai, Xiaolin and Liang, Yanyan and Chen, Min},
  booktitle = {Advances in Neural Information Processing Systems (NeurIPS) 2024},
  year      = {2024},
  url       = {https://openreview.net/forum?id=OoOCoZFVK3},
  urldate   = {2025-07-28},
  language  = {en}
}

@article{brodersen2015inferring,
  title={Inferring causal impact using Bayesian structural time-series models},
  author={Brodersen, Kay H and Gallusser, Fabian and Koehler, Jim and Remy, Nicolas and Scott, Steven L},
  year={2015}
}

@article{malecka2020application,
  title={Application of genetic algorithm to optimal income taxation},
  author={Ma{\l}ecka-Ziembi{\'n}ska, Edyta and Ziembi{\'n}ski, Rados{\l}aw},
  journal={Journal of Risk and Financial Management},
  volume={13},
  number={11},
  pages={251},
  year={2020},
  publisher={MDPI}
}

@inproceedings{
yao2023react,
title={ReAct: Synergizing Reasoning and Acting in Language Models},
author={Shunyu Yao and Jeffrey Zhao and Dian Yu and Nan Du and Izhak Shafran and Karthik R Narasimhan and Yuan Cao},
booktitle={The Eleventh International Conference on Learning Representations },
year={2023},
url={https://openreview.net/forum?id=WE_vluYUL-X}
}

@article{shinn2023reflexion,
  title={Reflexion: Language agents with verbal reinforcement learning},
  author={Shinn, Noah and Cassano, Federico and Gopinath, Ashwin and Narasimhan, Karthik and Yao, Shunyu},
  journal={Advances in neural information processing systems},
  volume={36},
  pages={8634--8652},
  year={2023}
}

@article{wei2022chain,
  title={Chain-of-thought prompting elicits reasoning in large language models},
  author={Wei, Jason and Wang, Xuezhi and Schuurmans, Dale and Bosma, Maarten and Xia, Fei and Chi, Ed and Le, Quoc V and Zhou, Denny and others},
  journal={Advances in neural information processing systems},
  volume={35},
  pages={24824--24837},
  year={2022}
}

@article{xu2023language,
  title={Language agents with reinforcement learning for strategic play in the werewolf game},
  author={Xu, Zelai and Yu, Chao and Fang, Fei and Wang, Yu and Wu, Yi},
  journal={arXiv preprint arXiv:2310.18940},
  year={2023}
}

@article{lowe2017multi,
  title={Multi-agent actor-critic for mixed cooperative-competitive environments},
  author={Lowe, Ryan and Wu, Yi I and Tamar, Aviv and Harb, Jean and Pieter Abbeel, OpenAI and Mordatch, Igor},
  journal={Advances in neural information processing systems},
  volume={30},
  year={2017}
}
\clearpage

\appendix

\section*{Appendix}
\addcontentsline{toc}{section}{Appendix}

\vspace{0.3em}
\noindent\textbf{Appendix Table of Contents}
\vspace{0.3em}

\begin{flushleft}
\renewcommand{\arraystretch}{1.1}
\begin{tabular}{@{}ll}
  \textbf{A.} & Representative Text Examples (Section~\ref{app:text_examples}) \\
  \textbf{B.} & Prompt Template           (Section~\ref{app:prompt_template}) \\
  \textbf{C.} & Experimental Setup and Results (Section~\ref{app:setup}) \\
\end{tabular}
\end{flushleft}

\vspace{0.8em}
\hrule
\vspace{1em}

\section{Representative Text Examples}
\label{app:text_examples}
\begin{tcolorbox}[title=\textbf{Short-term}, breakable]
\small

\begin{description}[leftmargin=0pt, labelsep=0.6em, style=nextline, font=\ttfamily]

  \item[Example 1]
  \textbf{Reasoning:} ``The current economic conditions are characterized by significant economic disparity, with both the top 10\% and bottom 50\% experiencing severe declines in wealth and income. The surge in wage rates has not translated into proportional gains in productivity or income, indicating a fragile economic environment. The family's personal productivity (0.7741156555876485) and wealth (0.09572083033270215) place them in a vulnerable position, with low wealth levels and moderate productivity. Given the economic volatility and the potential for increased social unrest and economic instability, the economic status is rated as 'Bad'.''\\
  \textbf{Economic status:} \texttt{0}\\

  \item[Example 2]
  \textbf{Reasoning:} ``The recent economic data indicate a significant improvement in wealth and income for the bottom 50\% of the population, with wealth increasing by 259.02\% and income by 221.43\%. This surge is accompanied by a notable boost in productivity (90.87\%) and an overall wage rate increase of 15.34\%. These positive trends suggest a more equitable distribution of economic gains, which can drive higher consumer spending and economic growth. The family's personal productivity (6.79) and wealth (3.48) are above average, placing them in a favorable position to benefit from these economic improvements. While maintaining a balance between increased labor time and managed consumption is crucial, the current economic conditions are robust and supportive, warranting a 'Good' rating.''\\
  \textbf{Economic status:} \texttt{2}\\

\end{description}

\end{tcolorbox}
\begin{tcolorbox}[title=\textbf{Long-term}, breakable]
\small

\begin{description}[leftmargin=0pt, labelsep=0.6em, style=nextline, font=\ttfamily]

  \item[Example 1]
  \textbf{Statement:}``We should advocate for policies that promote fair wage growth and equitable wealth distribution to stabilize the broader economic environment and benefit all families.''\\
  \textbf{Reasoning:}``The economic data reveal significant volatility and disparities in wealth, income, and productivity between the top 10\% and the bottom 50\% of the population. The top 10\% experience substantial fluctuations in wealth and income, while the bottom 50\% face more erratic and often negative changes. Despite some positive trends in productivity and wage rates, the overall economic environment remains challenging, with a low wage rate and persistent wealth gaps. The family's personal productivity and wealth place them in a moderate but vulnerable position, requiring careful financial management and strategic decisions to improve economic stability and self-utility.Given the neutral economic conditions, the family should focus on maintaining a balance between productivity and personal well-being. Overworking can reduce utility, while prudent consumption and savings can improve financial stability. Investing in education and skill development can enhance productivity and long-term economic resilience. Advocating for fair economic policies, such as progressive taxation and minimum wage increases, can help address broader inequalities and create a more stable economic environment, benefiting the family in the long run.''\\
  \textbf{Reflection:} ``The consensus among the households emphasizes the importance of balancing increased labor time with managed consumption to maintain personal well-being and financial stability. This aligns with our internal reasoning, which highlights the need for careful financial management and strategic decisions. Given our moderate but vulnerable economic position, we should focus on maintaining a balance to avoid overworking, which can reduce utility, while ensuring prudent consumption and savings. Investing in education and skill development will enhance productivity and long-term economic resilience. We should also advocate for fair economic policies to address broader inequalities and create a more stable environment for our family.''
  
  \textbf{Economic status:} \texttt{1}\\
  \textbf{Belief:} \texttt{[0, 1, 0, 0, 0, 1, 1, 1, 0, 2]}\\
  \textbf{Trust:} \texttt{[8, 9, 9, 8, 9, 8, 9, 9, 8, 10]}
  
  \item[Example 2]
  \textbf{Statement:} ``We urge policymakers to implement progressive taxation and minimum wage increases to address the growing wealth and income disparities and promote a more equitable economic environment."\\
  \textbf{Reasoning:} ``The economic data reveal significant volatility and disparities in wealth, income, and productivity between the top 10\% and the bottom 50\% of the population. The top 10\% experience substantial fluctuations in wealth and income, while the bottom 50\% face more erratic and often negative changes. Despite some positive trends in productivity and wage rates, the overall economic environment remains challenging, with a low wage rate and persistent wealth gaps. The family's personal productivity and wealth place them in a moderate but vulnerable position, requiring careful financial management and strategic decisions to improve economic stability and self-utility.Given the neutral economic conditions, the family should focus on maintaining a balance between productivity and personal well-being. Overworking can reduce utility, while prudent consumption and savings can improve financial stability. Investing in education and skill development can enhance productivity and long-term economic resilience. Advocating for fair economic policies, such as progressive taxation and minimum wage increases, can help address broader inequalities and create a more stable economic environment, benefiting the family in the long run.''\\
  \textbf{Reflection:} ``The statements from other households highlight the importance of balancing increased labor time with managed consumption to maintain personal well-being and financial stability. Given our moderate but vulnerable economic position, it is crucial to avoid overworking, which can reduce utility, and instead focus on prudent financial management. Investing in education and skill development can enhance productivity and long-term economic resilience. Advocating for fair economic policies can also help address broader inequalities and create a more stable environment, benefiting our family in the long run.''\\

  \textbf{Economic status:} \texttt{1}\\
  \textbf{Belief:} \texttt{[1, 2, 0, 0, 0, 1, 1, 0, 0, 1]}\\
  \textbf{Trust:} \texttt{[8, 7, 8, 8, 9, 8, 9, 8, 8, 6]}
\end{description}

\end{tcolorbox}

\section{Prompt Template}
\label{app:prompt_template}

\begin{tcolorbox}[title=\textbf{Long‐term reasoning},breakable]
\small

You are a family decision inferent. Analyze the given data and provide insights.

Long‐Term News: \{long\_term\_news\}

Private Observation:
\begin{itemize}
  \item Personal productivity (e): \{private\_observation[0]\}
  \item Personal wealth: \{private\_observation[1]\}
\end{itemize}

Similar Experiences: \{similar\_experience if similar\_experience else "No similar experiences found."\}

Your final goal is to improve the self‐utility of the current family, where increased labor time reduces utility and increased consumption improves utility, under the Bewley–Aiyagari model.

\textbf{Tasks:}
\begin{enumerate}
  \item Summarize key economic insights in “analysis”.
  \item Rate the economic condition as:
    \begin{itemize}
      \item 0 = Bad
      \item 1 = Neutral
      \item 2 = Good
    \end{itemize}
    Store this as “economic\_status”.
  \item Based on the current situation and private observation, give suggestions in “reasoning”.
  \item Generate 3 unique public statements in “statements”.
\end{enumerate}

Return exactly this JSON (no extra keys or commentary):
\begin{verbatim}
{
  "analysis": "...",
  "economic_status": 0,
  "reasoning": "...",
  "statements": ["...", "...", "..."]
}
\end{verbatim}

\end{tcolorbox}

 \begin{tcolorbox}[title=\textbf{Short-term reasoning},breakable]
\small
You are a family decision inferent. Your goal is to improve the family’s self-utility under the Bewley–Aiyagari model (more labor ↓ utility, more consumption ↑ utility).

\textbf{Inputs:}
\begin{itemize}
  \item Short-Term News: \{short\_term\_news\}
  \item Recent Long-Term News: \{recent\_long\_term\_result if recent\_long\_term\_result else "None"\}
  \item Private Observation:
    \begin{itemize}
      \item Personal productivity (e): \{private\_observation[0]\}
      \item Personal wealth: \{private\_observation[1]\}
    \end{itemize}
\end{itemize}

\textbf{Tasks:}
\begin{enumerate}
  \item Provide a detailed analysis of current economic conditions and give actionable guidance for choosing savings rate and working hours.
  \item Rate the economic condition:
    \begin{itemize}
      \item 0 = Bad
      \item 1 = Neutral
      \item 2 = Good
    \end{itemize}
\end{enumerate}

\textbf{Output:}  
Return exactly this JSON (no extra keys or commentary):
\begin{verbatim}
{
  "economic_status": 0,
  "reasoning": "..."
}
\end{verbatim}
\end{tcolorbox}

\begin{tcolorbox}[title=\textbf{Reflection and update belief},breakable]
\small
You are a family decision inferent. Analyze the given other households’ statements and provide private insights.

Private Observation:
\begin{itemize}
  \item Personal productivity (e): \{private\_observation[0]\}
  \item Personal wealth: \{private\_observation[1]\}
\end{itemize}

Internal Reasoning: \{personal\_reasoning\}

Public Personal Statement: \{personal\_statement\}

Other Households’ Statements: 
\{chr(10).join([f"- {stmt}" for stmt in other\_agents\_statements])\}

Your final goal is to improve the self‐utility of the current family, where increased labor time reduces utility and increased consumption improves utility, under the Bewley–Aiyagari model.

\textbf{Tasks:}
\begin{enumerate}
  \item Classify each household’s wealth level as \texttt{wealth\_guesses} (0=Low, 1=Medium, 2=High) with exactly \{expected\_num\} elements. Notice one has status 2, four have status 1, and five have status 0.
  \item Rate each statement’s trustworthiness from 0 (not trustworthy) to 10 (highly trustworthy) as \texttt{trust\_levels} with exactly \{expected\_num\} elements.
  \item Provide a brief reflection in \texttt{reflection\_text}, focusing on yourself, others’ statements, and ensuing economic decisions.
\end{enumerate}

Return exactly this JSON (no extra keys or commentary):
\begin{verbatim}
{
  "wealth_guesses": [...],
  "trust_levels": [...],
  "reflection_text": "..."
}
\end{verbatim}
\end{tcolorbox}

\section{Experimental Setup and Results}
\label{app:setup}

In this appendix, we provide further details on our LAMP framework’s methodology (Appendix~\ref{app:method}) and experimental setup and results (Appendix~\ref{app:experience}). We elaborate on the mathematical formulations, training procedure, and environment configurations that were summarized in the main text. We also include additional results and explanations, including detailed scenario parameters and expanded discussions of Economic Slowdown (S2) and Crisis Shock (S3) from the main paper.

\subsection{Method}\label{app:method}

\textbf{Economic Environment and Tax Functions.} Our multi-agent economic environment (TaxAI) is based on a heterogeneous-agent macroeconomic model with a government and $N_h$ households. In each period, the government sets five policy variables: labor income tax $(\tau_t, \xi_t)$, wealth tax $(\tau_{a,t}, \xi_{a,t})$, and a public spending ratio $r^G_t = G_t / Y_t$. Here $\tau$ and $\tau_a$ control the average tax rates, while $\xi$ and $\xi_a$ control the progressivity (marginal rate) of the income and wealth taxes. 
The income and asset tax functions follow nonlinear HSV formulations:
\[
T(i_t) = i_t - (1-\tau)\frac{i_t^{1-\xi}}{1-\xi},\;
T^a(a_t) = a_t - \frac{1 - \tau_a}{1 - \xi_a} a_t^{1 - \xi_a}
\]
where \(T(\cdot)\) and \(T^a(\cdot)\) represent the income and asset tax schedules respectively, and \(\tau, \tau_a, \xi, \xi_a\) control the average and marginal tax rates.
The total tax revenue \(T_t\) is composed of income tax, wealth tax, and consumption tax across all households, \[T_t = \sum_{i=1}^N \bigl(T(i_t^i) + T(a_t^i) + \tau_s c_t^i\bigr)\].

\begin{table*}[t]
  \centering
  \small
  \begin{tabular}{lcccc}
    \toprule
    \textbf{Algorithms} & \textbf{Avg. Reward} & \textbf{Social Welfare} & \textbf{Consumption} & \textbf{Labor} \\
    \midrule
    \rowcolor{lampbg}\textbf{LAMP (Ours)}
      & \textbf{8.21 $\pm$ 0.12}
      & \textbf{2.10e+03 $\pm$ 6.93e+02}
      & 2.02e+05 $\pm$ 7.35e+04
      & 3.72e+05 $\pm$ 1.42e+05 \\
    \midrule
    \textbf{MADDPG}
      & 5.07 $\pm$ 0.16
      & 1.17e+03 $\pm$ 3.77e+02
      & 4.07e+05 $\pm$ 6.46e+04
      & 7.42e+05 $\pm$ 1.50e+05 \\
    \textbf{Rule-Based}
      & 7.39 $\pm$ 0.45
      & 1.88e+03 $\pm$ 6.09e+02
      & 2.65e+05 $\pm$ 6.35e+04
      & 6.15e+05 $\pm$ 1.20e+05 \\
    \textbf{Random}
      & 6.17 $\pm$ 0.41
      & 1.57e+03 $\pm$ 5.20e+02
      & 1.18e+05 $\pm$ 2.88e+04
      & 5.20e+05 $\pm$ 1.01e+05 \\
    \midrule
    \textbf{LLM-Only}
      & 5.94 $\pm$ 0.31
      & 1.78e+03 $\pm$ 9.36e+01
      & 2.45e+05 $\pm$ 4.71e+04
      & 1.02e+06 $\pm$ 2.16e+05 \\
    \textbf{CoT}
      & 6.33 $\pm$ 0.35
      & 1.90e+03 $\pm$ 1.06e+02
      & 3.42e+05 $\pm$ 8.59e+04
      & 1.03e+06 $\pm$ 2.20e+05 \\
    \textbf{ReAct}
      & 6.95 $\pm$ 0.22
      & 2.08e+03 $\pm$ 6.46e+01
      & 4.46e+05 $\pm$ 8.47e+04
      & 9.59e+05 $\pm$ 1.58e+05 \\
    \textbf{Reflexion}
      & 3.54 $\pm$ 0.40
      & 1.06e+03 $\pm$ 1.20e+02
      & 1.82e+05 $\pm$ 4.59e+04
      & 1.15e+06 $\pm$ 2.42e+05 \\
    \bottomrule
  \end{tabular}
  \caption{Performance comparison under Scenario S2 (Economic Slowdown).}
  \label{tab:scenario2_results}
\end{table*}

\begin{table*}[t]
  \centering
  \small
  \begin{tabular}{lcccc}
    \toprule
    \textbf{Algorithms} & \textbf{Avg. Reward} & \textbf{Social Welfare} & \textbf{Consumption} & \textbf{Labor} \\
    \midrule
    \rowcolor{lampbg}\textbf{LAMP (Ours)}
      & \textbf{8.18 $\pm$ 0.16}
      & \textbf{2.33e+03 $\pm$ 3.16e+02}
      & 1.96e+05 $\pm$ 3.14e+04
      & 3.21e+05 $\pm$ 5.93e+04 \\
    \midrule
    \textbf{MADDPG}
      & 5.19 $\pm$ 0.34
      & 1.13e+03 $\pm$ 5.69e+02
      & 5.49e+05 $\pm$ 3.10e+05
      & 8.61e+05 $\pm$ 4.71e+05 \\
    \textbf{Rule-Based}
      & 7.36 $\pm$ 0.38
      & 2.09e+03 $\pm$ 2.53e+02
      & 2.71e+05 $\pm$ 4.92e+04
      & 5.95e+05 $\pm$ 1.02e+05 \\
    \textbf{Random}
      & 6.24 $\pm$ 0.29
      & 1.77e+03 $\pm$ 2.14e+02
      & 1.21e+05 $\pm$ 2.09e+04
      & 5.05e+05 $\pm$ 9.87e+04 \\
    \midrule
    \textbf{LLM-Only}
      & 6.03 $\pm$ 0.32
      & 1.81e+03 $\pm$ 9.68e+01
      & 2.73e+05 $\pm$ 5.47e+04
      & 1.02e+06 $\pm$ 2.22e+05 \\
    \textbf{CoT}
      & 6.46 $\pm$ 0.35
      & 1.94e+03 $\pm$ 1.05e+02
      & 3.92e+05 $\pm$ 1.02e+05
      & 1.03e+06 $\pm$ 2.20e+05 \\
    \textbf{ReAct}
      & 7.05 $\pm$ 0.18
      & 2.11e+03 $\pm$ 5.53e+01
      & 5.00e+05 $\pm$ 7.60e+04
      & 9.58e+05 $\pm$ 1.52e+05 \\
    \textbf{Reflexion}
      & 3.68 $\pm$ 0.41
      & 1.10e+03 $\pm$ 1.24e+02
      & 2.10e+05 $\pm$ 5.60e+04
      & 1.15e+06 $\pm$ 2.42e+05 \\
    \bottomrule
  \end{tabular}
  \caption{Performance comparison under Scenario S3 (Crisis Shock).}
  \label{tab:scenario3_results}
\end{table*}

\textbf{Think–Speak–Decide Pipeline Recap.} In the main text, we introduced the three core modules of LAMP: Think, Speak, and Decide. For completeness, we restate how these modules function and detail how their outputs are integrated into the learning process:

(1)Think Module: At specific times, the environment produces a natural-language news description of the state of the economy, which agents use for reasoning. To ensure agents focus on the appropriate temporal scale, we schedule two types of news events as described in the main text (Section 3). At fixed long-term intervals $L_i$ (e.g., every $L$ steps), a long-term news summary $\mathcal{R}^{long}_{L_i}$ is generated by an LLM based on the recent trajectory of global observations. This reflects structural trends (e.g. sustained growth slowdown or rising inequality over time). Meanwhile, at any intermediate step, if there is a sudden significant change in key indicators, a short-term news $\mathcal{R}^{short}_t$ is triggered to announce the shock. Formally, letting $\mathcal{X}_t = (G_w(t), W(t), Y(t))$ represent the current values of critical metrics (wealth Gini, social welfare, and per-capita GDP, respectively), we set a shock threshold $\sigma$. If $\max_j |\mathcal{X}_{j,t} - \mathcal{X}_{j,t-1}| > \sigma$ for any metric $j$, then $\text{type}(t) = \textit{short}$; if $t$ coincides with a long-term checkpoint $L_i$, then $\text{type}(t) = \textit{long}$; otherwise no news is issued ($\text{type}(t) = \textit{none}$). This mechanism, summarized by Equation (4) in the main paper, ensures that agents receive timely, context-rich language updates rather than raw numbers – similar to how real economic agents rely on news media for important developments. In our implementation, we chose $\sigma$ and $L$ so that long-term news arrives periodically (every few years of simulation) and short-term news flags large quarterly swings in indicators (exact values are chosen to balance frequency of news with not overwhelming the agent with constant messages). Given a news text, each household agent uses a large language model $\mathcal{L}_{reason}$ to interpret the news relative to its own state. The agent produces a short private reasoning $\psi_t^i$ which may include its assessment of the economy (e.g., “good” or “bad” times, encoded as an economic status label 2/1/0) and a rationale for its next action (e.g., “reduce consumption and save more because a recession is coming”). In generating this reasoning, the agent can draw upon an experience pool of past reasoning trajectories. We maintain two experience memories per agent: a short-term memory $\mathcal{H}^{short}_{t,i}$ that caches the agent’s top reasoning trajectories from recent steps, and a long-term memory $\mathcal{H}^{long}$ that indexes high-value reasoning trajectories from across all agents and past episodes using a FAISS similarity index. At the start of a long-term reasoning phase, each agent retrieves a few most relevant past experiences $\mathrm{kNN}_{k_3}(\mathcal{H}^{l})$ (based on similarity of current news and state to past situations) and combines them with its recent short-term experiences $\mathcal{H}^{s}_{t,i}$ as contextual examples for the LLM prompt. This helps the agent “remember” successful strategies or important lessons from history, improving stability in sparse-reward, long-horizon settings. After the LLM produces the new reasoning $\psi_t^i$, we store the trajectory and its outcome (e.g., obtained reward) back into the short-term memory, and periodically (at long-term checkpoints) update the long-term memory with top trajectories from all agents. This design mitigates forgetting and allows re-use of good strategies, as evidenced by the performance drop when disabling the experience pool (see ablation results).

(2)Speak Module: After forming its private reasoning, each agent may broadcast a concise public message summarizing its strategy or perspective. To generate a message, we use another LLM $\mathcal{L}_{stmt}$ that takes as input the agent’s state and reasoning and produces a few candidate statements. An internal scoring function (a self-attention mechanism) selects one statement $v_t^i$ to broadcast. At a long-term news step (when agents typically communicate strategic intent), all agents exchange these statements simultaneously, resulting in a set $V_t = \{v_t^1, v_t^2, \dots, v_t^{N_h}\}$ visible to everyone. Each agent then interprets the incoming messages using a reflection function $\mathcal{L}_{reflect}$. This produces: (a) an updated belief about each other agent’s hidden state (for example, agent $i$ may infer whether agent $j$ is likely wealthy or poor based on $j$’s message, denoted $w_t^{i\to j} \in \{\text{low, mid, high}\}$), (b) a trust score $\tau_t^{i\to j} \in [0,10]$ indicating how credible or relevant agent $j$’s message is according to $i$, and (c) a short self-reflection $\alpha_t^i$ where agent $i$ articulates any revised understanding of its own situation after hearing others (e.g., “others are optimistic about the market, perhaps I should not be too conservative”). These reflection outputs effectively let agents do opponent modeling and belief updates via language. They are fed back into the Think module in the next cycle (closing the reasoning–communication loop) and also incorporated into the policy’s state input for decision-making. In summary, the Speak module enables strategic communication that improves coordination and adaptability: it ensures each agent is not reasoning in isolation, but rather adjusting its policy in light of peers’ stated intentions and perceived credibility.

(3) Decide Module: The Decide stage integrates the numerical and language information to output final actions through a reinforcement learning policy. We use a centralized training, decentralized execution (CTDE) paradigm with an actor–critic algorithm (based on MADDPG). Specifically, during training, a centralized critic \(Q_{\phi}(x, a^1,\dots,a^{N_h}, a^{\mathrm{gov}})\) takes as input the joint state and joint action of all agents, and outputs a Q-value (expected cumulative reward) to critique the action choices. For households, the actors are decentralized but group-shared policies
\(
a_t^i = \mu_{\theta_{g(i)}}(o_t^i, m_t^i),
\)
where \(g(i)\in\{1,2,3\}\) denotes the wealth group of household \(i\) (top 10\%, middle 40\%, bottom 50\%). The government uses a separate actor
\(
a_t^{\mathrm{g}}=\mu_{\theta_{\mathrm{g}}}(O_t^g).
\)
Each household policy observes the agent’s execution-time observation \(o_t^i\), consisting of the shared global observation \(O_t^g\) and its private numerical state \(o_{t}^{i,h}\), augmented with its own language-based context \(m_t^i\). Here \(m_t^i\) is a fixed-size vector representation of textual inputs relevant to agent \(i\) at time \(t\), including its private reasoning \(\psi_t^i\) and its reflection \(\alpha_t^i\) (concatenated or pooled). To obtain \(m_t^i\), we encode the texts with a pretrained language encoder \(E_{\text{text}}\) and project them to a lower dimension \(d\) using a linear layer \(P(\cdot)\). This way, the language information enters the policy network in a controlled, compact form rather than raw text tokens, which greatly improves learning efficiency. The critic state \(x_t\) at time \(t\) consists of the shared global numerical observation, all households’ private numerical observations, and all households’ language embeddings:
\(
x_t = \bigl(O_t^g,\; o_{t}^{\,1:\!N_h},\; m_t^{\,1:\!N_h}\bigr).
\)
The critic uses this state to evaluate joint actions. We train the critic by minimizing the mean squared Bellman error:
\[
L_{\text{critic}} =
  \mathbb{E}_{(\,x_t,a_t,r_t,x_{t+1})\sim\mathcal D}
  \Bigl[
    \bigl(Q_{\phi}(x_t,a_t)-y_t\bigr)^2
  \Bigr]
\]
with the target value
\[
y_t =
  r_t + \gamma\,
  Q_{\phi'}\bigl(x_{t+1},\,a'_{t+1}\bigr),
\]
where \(a'_{t+1}\) is the joint next action produced by the target household-group actors and the target government actor, and \(\phi'\) and \(\mu'\) denote target networks updated by Polyak averaging. For a household in group \(g(i)\), the shared actor aims to maximize the expected return
\[
J(\theta_{g(i)}) =
\mathbb{E}_{\mathcal D}\Bigl[
  Q_{\phi}\bigl(
    x_t,\,
    a_{-i},\,
    \mu_{\theta_{g(i)}}(o_t^{\,i},m_t^{\,i})
  \bigr)
\Bigr].
\]
In practice, we minimize the negative-\(Q\) actor loss
\[
L_{\text{actor}}(\theta_{g(i)}) =
- \mathbb{E}_{\mathcal D}\Bigl[
  Q_{\phi}\bigl(
    x_t,\,
    a_{-i},\,
    \mu_{\theta_{g(i)}}(o_t^{\,i},m_t^{\,i})
  \bigr)
\Bigr],
\]
so that gradient descent on \(L_{\text{actor}}\) is equivalent to gradient ascent on \(J(\theta_{g(i)})\). This setup makes language an explicit, compact control signal via encoder–projection compression, rather than mere raw text concatenation.

\begin{table*}[t]
  \centering
  \small
  \setlength{\tabcolsep}{6pt}
  \begin{tabular}{lcccc}
    \toprule
    \textbf{Scenario} & \textbf{Depreciation Rate} & \textbf{Consumption Tax Rate} & \textbf{Interest Rate} \\
    \midrule
    S1: Economic Stability
      & 0.06
      & 0.065
      & 0.04 \\
    S2: Economic Slowdown
      & 0.12
      & 0.02
      & 0.08 \\
    S3: Crisis Shock
      & 0.10
      & 0.10
      & 0.10 \\
    \bottomrule
  \end{tabular}
  \caption{Hyperparameter settings for the three economic scenarios (S1–S3).}
  \label{tab:scenario_params}
\end{table*}

\begin{table}[t]
  \centering
  \small
  \setlength{\tabcolsep}{6pt}
  \begin{tabular}{lc}
    \toprule
    \textbf{Model} & \textbf{Avg. Reward} \\
    \midrule
    DeepSeek-v3.1 & 8.64 \\
    Qwen3-32B     & 8.35 \\
    Gemini-2.5    & 8.65 \\
    \bottomrule
  \end{tabular}
  \caption{Average reward of LAMP with different LLM backbones.}
  \label{tab:llm_backbones}
\end{table}

\begin{figure*}[t]
\centering
\includegraphics[width=1\textwidth]{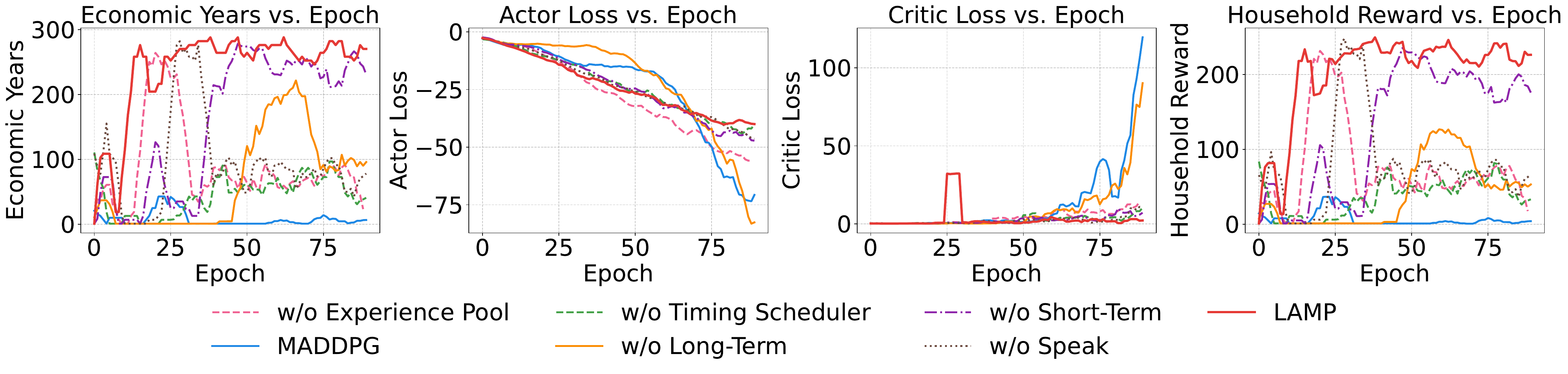} 
 \vskip -0.1 in 
\caption{Training curves over the first 80 epochs for seven methods: Economic Years, Actor Loss, Critic Loss, and Household Reward. LAMP (solid line) converges to higher and more stable values, with faster and smoother loss reduction and the highest household rewards, highlighting its advantage over baselines and ablation variants.}
\label{fig:training_curves}
 \vskip -0.1 in 
\end{figure*}

\subsection{Experimental Setup and Additional Results}\label{app:experience}

\textbf{Environment Scenarios.} We evaluate LAMP and baseline methods in three distinct economic scenarios, all simulated in the TaxAI environment described above. Each scenario corresponds to a different setting of structural parameters to mimic various macroeconomic conditions:

S1: Baseline Economic Stability. This scenario uses standard calibrated parameters intended to reflect a stable, growing economy. For instance, the annual capital depreciation rate is set to 6\%, the consumption tax rate is 6.5\%, and the nominal interest rate is 4\%. This scenario was used to train the agents and represents normal conditions without major external shocks.

S2: Economic Slowdown. In this scenario, we introduce a moderate supply and demand shift to simulate a slowdown or mild recession. We double the depreciation rate to 12\% (0.12) – meaning capital assets lose value faster, modeling a slump in productivity or faster obsolescence. To counteract weaker demand, the consumption tax rate is lowered to 2\% (down from 6.5\%), representing a fiscal stimulus to encourage spending. Meanwhile, we raise the interest rate to 8\% (0.08), reflecting tighter credit conditions or an anti-inflationary stance by the monetary authority during the slowdown. These changes result in generally tougher conditions for growth: capital accumulation is harder (due to high depreciation and interest), although consumers get a tax break. We expect agents to adapt by, e.g., saving less (since returns are lower) and working a bit more to maintain income.

\begin{table}[t]
  \centering
  \small
  \setlength{\tabcolsep}{6pt}
  \begin{tabular}{lcl}
    \toprule
    \textbf{Hyperparameter} & \textbf{Value} & \textbf{Algorithm / Module} \\
    \midrule
    $q_{\text{lr}}$           & 3e-4   & MADDPG (critic learning rate) \\
    $p_{\text{lr}}$           & 3e-4   & MADDPG (actor learning rate)  \\
    \texttt{buffer\_size}     & 1e6    & MADDPG (replay buffer)        \\
    $\gamma$                  & 9.75e-1& MADDPG (discount factor)      \\
    $\tau$                    & 5e-3   & MADDPG (target network update)\\
    \texttt{embed\_dim}       & 5e0    &  (language embedding size)\\
    \texttt{threshold}        & 4e-1   &  (shock detection)        \\
    \bottomrule
  \end{tabular}
  \caption{Key hyperparameters and their corresponding algorithm components or modules.}
  \label{tab:key_hyperparams}
\end{table}

S3: Crisis Shock. This scenario models a severe economic crisis with coupled shocks. We set a high consumption tax rate of 10\%, both to simulate increased fiscal burden (governments raising taxes in a crisis) and to represent high effective prices dampening consumption. The interest rate is also raised to 10\%, indicating very tight monetary conditions (e.g., a central bank fighting inflation or risk). The depreciation rate is set to 10\%, moderately higher than baseline (though slightly lower than S2’s 12\%, it still represents a significant supply shock where capital wears out quickly). 

For all scenarios, we simulate up to 300 periods (years) or until the economy “collapses” (e.g., if the environment diverges or a policy leads to an infeasible state).We use identical initial conditions across methods for fairness and run multiple random seeds (8) to account for stochasticity in learning and LLM generation.

\textbf{Additional Results}
Tables~\ref{tab:scenario2_results} and~\ref{tab:scenario3_results} report the key metrics—Average Household Reward, Social Welfare, Consumption, and Labor—of LAMP and seven baselines under Scenario S2 (Economic Slowdown) and Scenario S3 (Crisis Shock), respectively. In both settings, LAMP achieves the highest welfare and reward while maintaining competitive consumption and labor levels, demonstrating its robustness to macroeconomic shifts.

Beyond baseline comparisons, we further replace MADDPG with alternative non-language MARL algorithms and report the resulting average household rewards under the same real-data–calibrated economy. For MAPPO, LAMP attains an average reward of 8.67 compared to 8.61 for the numeric baseline. LAMP thus consistently matches or slightly outperforms these stronger numeric baselines, indicating that its gains are not tied to a particular MARL backbone. Table~\ref{tab:llm_backbones} varies the LLM backbone (DeepSeek-v3.1, Qwen3-32B, Gemini-2.5) while keeping the rest of LAMP unchanged. The average rewards remain similar across models, suggesting that LAMP’s benefits are robust to reasonable changes in the underlying language model.

\paragraph{Key Hyperparameters Summary}
Table~\ref{tab:key_hyperparams} lists the principal hyperparameters from our training configuration, indicating which algorithm or module each pertains to. Hyperparameters such as $q_{\text{lr}}$, $p_{\text{lr}}$, \texttt{buffer\_size}, $\gamma$, and $\tau$ govern the MADDPG training dynamics. The entropy coefficient (\texttt{ent\_coef}) and value-loss coefficient (\texttt{vloss\_coef}) are relevant in soft actor–critic and general actor–critic frameworks. The embedding dimension (\texttt{embed\_dim}) and shock threshold (\texttt{threshold}) are specific to the LAMP architecture’s language processing and Think module.

Except for MADDPG, which was trained for 200 epochs to ensure stable evaluation behavior and to prevent the environment from collapsing after a single interaction step, all other methods were trained for 80 epochs. 

\textbf{Analysis of Training Curves for LAMP and Baselines}
As shown in Figure~\ref{fig:training_curves}, the four panels plot key metrics over the first 80 training epochs for seven methods. In the top-left panel, LAMP’s solid line converges to a higher, more stable “Economic Years” value, indicating prolonged system stability. The top-right and bottom-left panels show that its Actor and Critic Loss curves decline more rapidly and with reduced oscillation, reflecting more efficient policy and value learning. Finally, in the bottom-right panel, LAMP achieves the highest and smoothest Household Reward, demonstrating its superior balance of labor and consumption under the same training budget. Overall, these curves underscore the effectiveness of the language-augmented LAMP framework in multi-agent economic simulations.

\end{document}